\documentclass{article}

\usepackage{arxiv}

\usepackage[utf8]{inputenc} 
\usepackage[T1]{fontenc}    
\usepackage{hyperref}       
\usepackage{url}            
\usepackage{booktabs}       
\usepackage{amsfonts}       
\usepackage{nicefrac}       
\usepackage{microtype}      
\usepackage{lipsum}		
\usepackage{graphicx}
\usepackage{natbib}
\usepackage{doi}
\usepackage[nolist,nohyperlinks]{acronym}
\usepackage[bottom]{footmisc}

\usepackage{amssymb}
\usepackage{amsmath}
\usepackage{latexsym}
\usepackage[ruled,vlined]{algorithm2e}
\usepackage{url}
\usepackage{xcolor}
\usepackage{dcolumn} 
\usepackage{hyperref}
\usepackage{threeparttable}
\usepackage{booktabs}
\usepackage{caption}
\usepackage{subcaption}

\begin{acronym}
    \acro{CRC}{Colorectal cancer}
    \acro{WSI}{whole slide image}
    \acro{HE}[H\&E]{Hematoxylin and Eosin}
    \acro{ReLU}{rectified linear activation unit}
    \acro{SRA}{Self-Rule to Adapt}
    \acro{SRMA}[SRMA]{Self-Rule to Multi-Adapt}
    \acro{CL}{contrastive learning}
    \acro{TMA}{tissue microarray}
    \acro{SOTA}{state-of-the-art}
    \acro{K16}{Kather-16}
    \acro{K19}{Kather-19}
    \acro{CRCTP}[CRC-TP]{Colorectal Cancer Tissue Phenotyping}
    \acro{E2H}{easy-to-hard}
    \acro{UDA}{unsupervised domain adaptation}
    \acro{ROIs}{regions of interest}
    \acro{SGD}{stochastic gradient descent}
    \acro{IoU}{intersection over union}
    \acro{TCGA}{The Cancer Genome Atlas}
    \acro{MIL}{Multiple-Instance Learning}
\end{acronym}

\title{Self-Rule to Multi-Adapt: Generalized Multi-source Feature Learning Using Unsupervised Domain Adaptation for Colorectal Cancer Tissue Detection}

\author{ \href{https://orcid.org/0000-0002-2879-9745}{\includegraphics[scale=0.06]{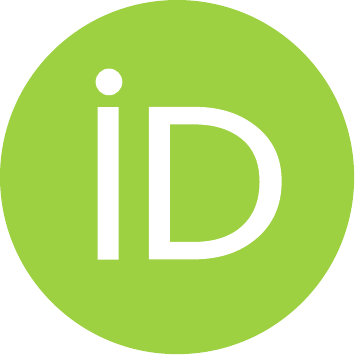}\hspace{1mm}Christian Abbet}\thanks{Co-first author. Christian Abbet and Linda Studer contributed equally.} \\
	Signal Processing Lab 5 (LTS5)\\
	EPFL\\
	Switzerland\\
	\texttt{christian.abbet@epfl.ch} \\
	\And
	Linda Studer${}^{*}$\\
	Documents, Image and Video Analysis (DIVA)\\
	University of Fribourg\\
	Switzerland\\
	\texttt{linda.studer@unifr.ch} \\
	\And
	Andreas Fischer\\
	Documents, Image and Video Analysis (DIVA)\\
	University of Fribourg\\
	Switzerland\\
	\And
	Heather Dawson \\
	Institute of Pathology\\
	University of Bern\\
	Switzerland\\
	\And
	\href{https://orcid.org/0000-0001-6741-3000}{\includegraphics[scale=0.06]{orcid.pdf}\hspace{1mm}Inti Zlobec} \\
	Institute of Pathology\\
	University of Bern\\
	Switzerland\\
	\And
	\href{https://orcid.org/0000-0002-5759-4896}{\includegraphics[scale=0.06]{orcid.pdf}\hspace{1mm}Behzad Bozorgtabar} \\
	Signal Processing Lab 5 (LTS5)\\
	EPFL\\
	Switzerland\\
	\And
	\href{https://orcid.org/0000-0003-2938-9657}{\includegraphics[scale=0.06]{orcid.pdf}\hspace{1mm}Jean-Philippe Thian} \\
	Signal Processing Lab 5 (LTS5)\\
	EPFL\\
	Switzerland\\
}

\newcommand{\new}[1]{{#1}}

\renewcommand{\shorttitle}{Self-Rule to Multi-Adapt}

\hypersetup{
pdftitle={Self-Rule to Multi-Adapt: Generalized Multi-source Feature Learning Using Unsupervised Domain Adaptation for Colorectal Cancer Tissue Detection},
pdfsubject={-},
pdfauthor={Christian Abbet, Linda Studer, Andreas Fischer, Heather Dawson, Inti Zlobec, Behzad Bozorgtabar, Jean-Philippe Thiran},
pdfkeywords={Computational pathology, self-supervised learning, unsupervised domain adaptation, colorectal cancer},
}

\begin{document}
\maketitle

\begin{abstract}
    Supervised learning is constrained by the availability of labeled data, which are especially expensive to acquire in the field of digital pathology.
    Making use of open-source data for pre-training or using domain adaptation can be a way to overcome this issue.
    However, pre-trained networks often fail to generalize to new test domains that are not distributed identically due to tissue stainings, types, and textures variations.
    Additionally, current domain adaptation methods mainly rely on fully-labeled source datasets.
    In this work, we propose \ac{SRMA}, which takes advantage of self-supervised learning to perform domain adaptation, and removes the necessity of fully-labeled source datasets.
    \ac{SRMA} can effectively transfer the discriminative knowledge obtained from a few labeled source domain's data to a new target domain without requiring additional tissue annotations. 
    Our method harnesses both domains' structures by capturing visual similarity with intra-domain and cross-domain self-supervision. Moreover, we present a generalized formulation of our approach that allows the framework to learn from multiple source domains.
    We show that our proposed method outperforms baselines for domain adaptation of colorectal tissue type classification \new{in single and multi-source settings}, and further validate our approach on an in-house clinical cohort. The code and \new{trained} models are available open-source: \texttt{\href{https://github.com/christianabbet/SRA}{https://github.com/christianabbet/SRA}}.
\end{abstract}


\section{Introduction}
\label{sec:introduction}

\ac{CRC} is one of the most common cancers worldwide, and its understanding through computational pathology techniques can significantly improve the chances of effective treatment~\citep{geessink2019computer, smit2020role} by refining disease prognosis and assisting pathologists in their daily routine.
The data used in computational pathology most often consists of \ac{HE} stained \acp{WSI}~\citep{hegde2019similar,lu2020data} and \acp{TMA}~\citep{arvaniti2018automated, nguyen2021classification}

Although fully supervised deep learning models have been widely used for a variety of tasks, including tissue classification~\citep{kather2019predicting}
and semantic segmentation~\citep{qaiser2019fast, chan2019histosegnet}, in practice, it is time-consuming and expensive to obtain fully-labeled data as it involves expert pathologists. 
This hinders the applicability of supervised machine learning models to real-world scenarios. Weakly supervised learning is a less demanding approach that does not depend on large labeled cohorts. Examples of this approach applied to digital pathology include \acp{WSI} classification~\citep{tellez2018gigapixel, silva2021weglenet} and \ac{MIL} based on diagnostic reports~\citep{campanella2019clinical}.
However, these methods still need an adequate training set to initialize the learning process, limiting the gain that can be achieved from using unlabeled samples.

Self-supervised learning was proposed to address limitations linked to labeled data availability. 
It involves a training scheme where "\textit{the data creates its own supervision}"\citep{abbeel2020unsup} to learn rich features from structured unlabeled data.
Applications of this approach in computational pathology include multiple tasks such as survival analysis~\citep{abbet2020divide}, \acp{WSI} classification~\citep{li2020dual}, and image retrieval~\citep{gildenblat2019self}. 

Over the years, various large data banks have been made available online containing samples from a variety of organs~\citep{weinstein2013cancer, litjens20181399, veta2019predicting}, such as the colon and rectum~\citep{kather2016multi, coad2016, read2016, kather2019predicting}. This opens up possibilities for transfer learning and domain adaptation.
Yet, using these data banks to develop computational pathology-based models for real-world scenarios remains challenging because of the domain gap, as these images were created under different imaging scenarios.
A tissue sample's visual appearance can be heavily affected by the staining procedure~\citep{otalora2019staining}, the type of scanner used~\citep{cheng2019assessing}, or other artifacts such as folded tissues~\citep{komura2018machine}. 

To tackle this issue, color normalization techniques~\citep{macenko2009method, zanjani2018deep, anand2019fast} have been widely adopted. 
Nevertheless, these techniques solely rely on image color information, while the morphological structure of the tissue is not taken into account~\citep{Tam2016method, Zarella2017alternative}. This could lead to unpredictable results in the presence of substantial staining variations and dark staining due to densely clustered tumor cells.

Another field of research that aims to improve the classification of heterogeneous \acp{WSI} is \ac{UDA}. 
These methods work by learning from a rich source domain together with the label-free target domain to have a well-performing model on the target domain at inference time. \ac{UDA} allows models to include a large variety of constraints to match relevant morphological features across the source and target domains.

DANN~\citep{ganin2015unsupervised}, for example, uses gradient reversal layers to learn domain-invariant features. 
Self-Path~\citep{koohbanani2020self} combines the DANN approach with self-supervised auxiliary tasks. 
The selected tasks reflect the structure of the tissue and are assumed to improve the stability of the framework when working with histopathological images. 
Such auxiliary tasks include hematoxylin channel prediction, Jigsaw puzzle-solving, and magnification prediction. Another example is CycleGAN~\citep{zhu2017unpaired}, which takes advantage of adversarial learning to map images between the source and target domain cyclically.
However, adversarial approaches can fall short because they do not consider task-specific decision boundaries and only try to distinguish the features as either coming from the source or target domain~\citep{saito2018maximum}.

A further issue is that most \ac{UDA} methods consider fully-labeled source datasets~\citep{dou2019domain} for domain adaptation. However, digital pathology mainly relies on unlabeled or partly-labeled data as the acquisition of fully labeled cohorts is often unfeasible.
In addition, recent approaches tend to treat domain adaptation as a closed-set scenario~\citep{carlucci2019domain}, which assumes that all target samples belong to classes present in the source domain, even though this is often not the case in a real-world scenario.

To overcome this, OSDA~\citep{saito2018open} proposes an adversarial open-set domain adaptation approach, where the feature generator has the option to reject mistrusted or unknown target samples as an additional class. 
In another recent work, SSDA~\citep{xu2019self} uses self-supervised domain adaptation methods that combine auxiliary tasks, adversarial loss, and batch normalization calibration across the source and target domains.

\new{Another domain adaptation framework DANCE \citep{NEURIPS2020_bb7946e7} proposes a universal domain adaptation method to address arbitrary category shifts based on neighborhood clustering on the unlabeled target domain in a self-supervised way. Then, entropy-based optimization is utilized for feature alignment of known categories and unknown ones are rejected, based on their entropy. The recently proposed method SENTRY \citep{prabhu2021sentry} uses unsupervised domain adaptation based on selective entropy optimization, in which the target domain samples are selected based on their predictive consistency under a set of randomly augmented views. Then, SENTRY selectively optimizes the model's entropy on these samples based on their consistency to induce the domain alignment.}
Finally, some approaches take advantage of multiple source datasets to learn features that are discriminant under varying modalities. In~\cite{matsuura2020domain}, domain-agnostic features are generated by combining a domain discriminator as well as a hard clustering approach. 

In this work, we propose a label-efficient framework called \new{\acf{SRMA}} for tissue type recognition in histological images and attempt to overcome the issues mentioned above by combining self-supervised learning approaches with \ac{UDA}. 
We present an entropy-based approach that progressively learns domain invariant features, thus making our model more robust to class definition inconsistencies as well as the presence of unseen tissue classes when performing domain adaptation.
\new{\ac{SRMA}} is able to accurately identify tissue types in \ac{HE} stained images, which is an important step for many downstream tasks.
Our proposed method achieves this by using few labeled open-source datasets and unlabeled data which are abundant in digital pathology, thus reducing the annotation workload for pathologists.
We show that our method outperforms previous domain adaptation approaches in a few-label setting and highlight the potential use for clinical application in the diagnostics of \ac{CRC}.

This study is an extension of the work we presented at the Medical Imaging with Deep Learning (MIDL) 2021 conference~\citep{abbet2021selfrule}.
Here, we provide a more in-depth explanation and analysis of our proposed entropy-based \ac{E2H} learning strategy.
Additionally, we reformulate the entropy-based cross-domain matching used by the \ac{E2H} learning strategy which improves the prediction robustness when dealing with complex tissue structures.
Moreover, we also provide the generalization of \new{the previously proposed \ac{SRA} framework to multi-source domain adaptation by including an additional public dataset and performing further experiments to assess the model's performance. Thus, we name this improved framework \acf{SRMA}.}

\section{Methods}
\label{sec:methods}

In our unsupervised domain adaptation scenario, we have access to a small set of labeled data sampled from a source domain distribution and a set of unlabeled data from a target distribution. The goal is to learn a hypothesis function (for example, a classifier) on the source domain that provides a good generalization in the target domain. 

\begin{figure*}[ht]
    \centering
    \includegraphics[width=0.99\textwidth]{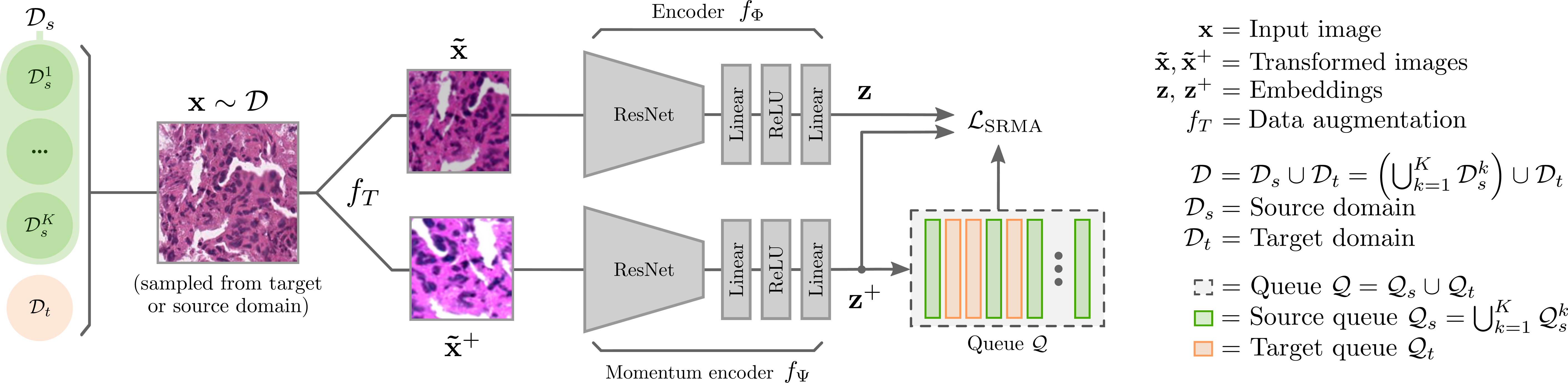}
    \caption{\new{Schematic overview of the \acf{SRMA} framework for a given input image $\mathbf{x}$ sampled from $\mathcal{D}=\mathcal{D}_s \cup \mathcal{D}_t = \bigcup_{k=1}^{K} \mathcal{D}_s^k \cup \mathcal{D}_t$. 
    Each encoder receives a different augmented version of the input image, generated by $f_{T}$.  
    The loss $\mathcal{L}_{\mathrm{SRMA}}=\mathcal{L}_{\mathrm{IND}}+\mathcal{L}_{\mathrm{CRD}}$ is the composition of the in-domain loss $\mathcal{L}_{\mathrm{IND}}$ and cross-domain loss $\mathcal{L}_{\mathrm{CRD}}$, which aims at reducing the domain gap between the source and target domains. The queue $\mathcal{Q}$ keeps track of previous samples' embeddings and their set of origin (source or target).}}
    \label{fig:model_pipeline_moco}
\end{figure*}

To this end, we propose a novel self-supervised cross-domain adaptation setting called \new{\ac{SRMA}}, which is described in more detail below. 
\new{We first introduce the architecture in a single-source setting and then present the generalization to the multi-source setting in Section \ref{subsec:generalization_multisource}}.
Figure~\ref{fig:model_pipeline_moco} gives an overview of the proposed framework, and Algorithm~\ref{alg:sra} presents the pseudo-code of our \ac{SRMA} method in the single-source setting.

To train our framework, we rely on a set of images $\mathcal{D} = \mathcal{D}_s \cup \mathcal{D}_t$ that is the aggregation of a set of source images $\mathcal{D}_s$ and a set of target images $\mathcal{D}_t$. 
The model takes as input an RGB image $\mathbf{x} \in \mathbb{R}^{H \times W \times 3}$ sampled from $\mathcal{D}$ where $H$ and $W$ denote the height and width of the image, respectively.
When sampling from $\mathcal{D}$, there is an equal probability to draw a sample from either the source or the target domain. 
After sampling, two sets of random transformations are applied to the image $\mathbf{x}$ using an image transformer $f_{T}: \mathbb{R}^{H \times W \times 3} \rightarrow \mathbb{R}^{H \times W \times 3}$. This generates a pair of augmented views $\tilde{\mathbf{x}}$, $\tilde{\mathbf{x}}^{+}\in \mathbb{R}^{H \times W \times 3}$ that are assumed to share similar content as they are both different augmentations of the same sampled input image.
Each image of the pair $\tilde{\mathbf{x}}$, $\tilde{\mathbf{x}}^{+}$ is then fed to its respective encoder $f_{\Phi} : \mathbb{R}^{H \times W \times 3} \rightarrow \mathbb{R}^d$ and $f_{\Psi} : \mathbb{R}^{H \times W \times 3} \rightarrow \mathbb{R}^d$ to compute the query $\mathbf{z} \in \mathbb{R}^{d}$ and key $\mathbf{z}^{+} \in \mathbb{R}^{d}$ embeddings of the input image. 
Here, $d$ represents the dimension of the embedding space. 
\new{For notational simplicity, when sampling an image $\mathbf{x}$, we directly assume its embedding as} $\mathbf{z}^{}, \mathbf{z}^{+} \in \mathcal{D}$.

Each network's branch consists of a residual encoder followed by two linear layers based on the \ac{SOTA} architecture proposed in~\cite{chen2020improved} (MoCoV2).
We use the key embeddings $\mathbf{z}^{+}$ to maintain a queue of negative samples $\mathcal{Q} = \{\mathbf{q}_l \in \mathbb{R}^d\}_{l=1}^{\lvert \mathcal{Q} \rvert}$ in a first-in, first-out fashion. 
When updating the queue with a new negative sample, not only the sampled image's embedding is stored but also its domain of origin (source or target). \new{It allows the architecture to know at anytime the domain of origin of each queue sample.}

The queue provides a large number of examples which alleviates the need for a large batch size \citep{chen2020simple} or the use of a memory bank \citep{kim2020cross}. \new{In addition, it enables the model to scale more easily as $\mathcal{D}$ grows as the size of the queue does not depend on it.} 
Moreover, $f_{\Psi}$ is updated using a momentum approach, combining its weights with those of $f_{\Phi}$.
This approach ensures that $f_{\Psi}$ generates a slowly shifting and, therefore, coherent embedding. 

Motivated by~\cite{ge2020self,kim2020cross, abbet2021selfrule}, we extend the domain adaptation learning procedure to our model definition and task. Hence, we split the loss terms into two distinct tasks, namely the in-domain $\mathcal{L}_{\mathrm{IND}}$ and cross-domain $\mathcal{L}_{\mathrm{CRD}}$ representation learning. The objective loss \new{$\mathcal{L}_{\mathrm{SRMA}}$} is the summation of both terms, which are described in more detail below.

\begin{equation}
    \new{\mathcal{L}_{\mathrm{SRMA}}} = \mathcal{L}_{\mathrm{IND}} + \mathcal{L}_{\mathrm{CRD}},
    \label{eq:l_sra}
\end{equation}

\subsection{In-domain Loss} 
The first objective $\mathcal{L}_{\mathrm{IND}}$ aims at learning the distribution of each the source and the target domain features individually. 
We want to keep the two domains independent as their alignment is optimized separately by the cross-domain loss term. 
For each embedding vector $\mathbf{z}$, there is a paired embedding vector $\mathbf{z}^+$ that is generated from the same sampled tissue image and therefore is, by definition, similar. As a result, their similarity can be jointly optimized using a contrastive learning approach \citep{oord2018representation}. Here, we strongly benefit from data augmentation to create discriminant features that match both $\mathbf{z}$ and $\mathbf{z}^+$, making them more robust to outliers. By selecting data augmentations suited to histology \new{\citep{tellez2019quantifying, faryna2021tailoring}, we can ensure that the learned features are consistent with naturally occurring data variations in histology, and therefore guide the model towards histopathologically meaningful representations.
This approach differs from \cite{kim2020cross}, where a memory bank is used instead of the combination of a queue and data augmentation to keep track of past samples.}
Therefore, the in-domain loss, as expressed in Equations~\ref{eq:p_ind}-\ref{eq:L_ind} constrains the representation of the embedding space for each domain separately.

\begin{equation}
    p_{\,_{\mathrm{IND}}}(\mathbf{z}, \mathbf{z}^+, \mathcal{Q}) = 
     \dfrac{\exp(\mathbf{z}^\top \mathbf{z}^{+}/\tau)}{\exp(\mathbf{z}^\top \mathbf{z}^{+}/\tau) + \sum\limits_{\mathbf{q}_l \in \mathcal{Q}} \exp(\mathbf{z}^{\top} \mathbf{q}_l/\tau)}.
    \label{eq:p_ind}
\end{equation}

\begin{equation}
    \new{l_{\,_{\mathrm{IND}}}(\mathcal{D}, \mathcal{Q}) = \sum_{\mathbf{z}^{\,}, \mathbf{z}^+ \in \mathcal{D}} \log{\left[ p_{\,_{\mathrm{IND}}}(\mathbf{z}^{\,}, \mathbf{z}^+, \mathcal{Q})\right]}.}
    \label{eq:l_ind}
\end{equation}

\begin{equation}
    \new{\mathcal{L}_{\mathrm{IND}} = \frac{-1}{\lvert \mathcal{D}_s \rvert +\lvert \mathcal{D}_t \rvert} \left[ l_{\,_{\mathrm{IND}}}(\mathcal{D}_s, \mathcal{Q}_s) + l_{\,_{\mathrm{IND}}}(\mathcal{D}_t, \mathcal{Q}_t) \right]}.
    \label{eq:L_ind}
\end{equation}

\begin{algorithm}[ht]
\SetAlgoLined
\small
\new{
Initialize queue $\mathcal{Q}$ by sampling from normal distribution $\mathcal{N}(0, 1)$\;
Normalize queue entries $\{q_i\} \in Q$\;
\For{$\mathrm{e} = 0$ \KwTo $N_\mathrm{epochs}-1$}{
    Create $\mathcal{D}$ by uniformly sampling from $\mathcal{D}_s$ and $\mathcal{D}_t^{}$\;
    Update easy-to-hard coefficient $r$ using Equation~\ref{eq:r}\;
    \For{$\mathrm{batch}$ $\{\mathbf{x}_i\}_{i=1}^B$ $\mathrm{in}$ $\mathcal{D}$}{
        Get augmented samples $\tilde{\mathbf{x}}_i^{\,}, \tilde{\mathbf{x}}^+_i$ using $f_T$\;
        Perform forward pass $\mathbf{z}_i^{\,}=f_{\phi}(\tilde{\mathbf{x}}_i^{\,}),\ \mathbf{z}^+_i= f_{\psi}(\tilde{\mathbf{x}}^+_i)$ \;
        Normalize vectors $\mathbf{z}_i^{\,}$, $\mathbf{z}^{+}_i$ \;
        Compute in-domain loss $\mathcal{L}_{\mathrm{IND}}$ using Equation~\ref{eq:L_ind}\;
        Calculate cross-entropy $\bar{H}$ using Equation~\ref{eq:h_mean_crd} \;
        Compute easy-to-hard $\mathcal{R}_{s}$, $\mathcal{R}_{t}^{}$ sets using Equation~\ref{eq:r_set} \;
        Evaluate cross-domain loss $\mathcal{L}_{\mathrm{CRD}}$ by replacing  $\mathcal{D}_{s}$, $\mathcal{D}_{t}$ with $\mathcal{R}_{s}$, $\mathcal{R}_{t}$ in Equation~\ref{eq:L_crd}, respectively\;
        Compute $\mathcal{L}_{\mathrm{SRA}} = \mathcal{L}_{\mathrm{IND}} + \mathcal{L}_{\mathrm{CRD}}$\;
        Update $f_{\Phi}$ weights with backpropagation \;
        Update $f_{\Psi}$ weights with momentum 
        \;
        Update queue $\mathcal{Q}$ with $\mathbf{z}^{+}_i$\;
    }
}
}
\caption{Pseudocode for the single-source \acf{SRMA} framework}
\label{alg:sra}
\end{algorithm}

We denote $\mathcal{Q}_{s}, \mathcal{Q}_{t} \subset \mathcal{Q}$ as the sets of indexed samples of the queue that were previously drawn from the corresponding domain $\mathcal{D}_{s}, \mathcal{D}_{t} \subset \mathcal{D}$, and $\tau \in \mathbb{R}$ as the temperature. The temperature is typically small ($\tau \ll 1$), thus sharpening the signal and helping the model to make confident predictions.
For all images of each dataset $\mathcal{D}_{s}, \mathcal{D}_{t}$, we want to minimize the distance between $\mathbf{z}$ and $\mathbf{z}^{+}$ while maximizing the distance to the previously generated negative samples from the corresponding sets $\mathcal{Q}_{s}, \mathcal{Q}_{t}$. The samples in the queue are considered reliable negative candidates as they are generated by $f_{\Psi}$ whose weights are slowly optimized due to its momentum update procedure.

\subsection{Cross-domain Loss} 
We can see the cross-domain matching task as the generation of features that are discriminative across both sets. 
In other words, two samples that are visually similar but are drawn from the source $\mathcal{D}_s$ and target $\mathcal{D}_t$ domain, respectively, should have a similar embedding. 
On the other hand, when comparing these samples 
to the remaining candidates of the opposite domain, their resulting embeddings should be far apart.
Practically, performing cross-domain matching using the number of available candidates within a batch might deteriorate the quality of the domain matching process due to the limited amount of negative samples.
Therefore, we use the queue to find negative samples for domain matching.
Hence, we compute the similarity and \new{cross-entropy} of each query pair $\mathbf{z}$, $\mathbf{z}^{+}$ drawn from one set (for example $\mathcal{D}_s$) to the stored queue samples from the other set (for example $\mathcal{Q}_t$): 

\begin{equation}
     p_{\,_{\mathrm{CRD}}}(\mathbf{z}, \mathbf{q}, \mathcal{Q}) = \frac{\exp(\mathbf{z}^\top \mathbf{q}/\tau)}{\sum\limits_{\mathbf{q_l} \in \mathcal{Q}} \exp(\mathbf{z}^\top \mathbf{q}_l/\tau)},
    \label{eq:ph_crd}
\end{equation}

\begin{equation}
    H(\mathbf{z}, \mathbf{z^+}, \mathcal{Q}) = - \sum_{\mathbf{q} \in \mathcal{Q}} p_{\,_{\mathrm{CRD}}}(\mathbf{z}, \mathbf{q}, \mathcal{Q}) \log{\left[p_{\,_{\mathrm{CRD}}}(\mathbf{z^+}, \mathbf{q}, \mathcal{Q})\right]},
    \label{eq:h_crd}
\end{equation}

\begin{equation}
    \bar{H}(\mathbf{z}, \mathbf{z^+}, \mathcal{Q})  = \frac{1}{2} \left[ H(\mathbf{z}^{\,}, \mathbf{z}^+, \mathcal{Q}) + H(\mathbf{z}^{+}, \mathbf{z}^{\,}, \mathcal{Q}) \right].
    \label{eq:h_mean_crd}
\end{equation}

A low \new{cross-entropy} $H$ means that the selected query pair $\mathbf{z}^{\,}$, $\mathbf{z}^+$ from one domain matches with a limited number of samples from another domain.
\new{Moreover, we update our initial definition of $H$ \citep{abbet2021selfrule}, where solely $\mathbf{z}$ is used.
By taking the average cross-entropy $\bar{H}$, the model is now also penalized when the predictions from $\mathbf{z}^{\,}$, $\mathbf{z}^+$ of the same image are different.
This improves the consistency of the domain matching\citep{assran2021semi}}.
\new{As a result, the loss $\mathcal{L}_{\textrm{CRD}}$ aims to minimize the averaged cross-entropy of the similarity distributions, assisting the model in making confident predictions}:

\begin{equation}
    \new{l_{\,_{\mathrm{CRD}}}(\mathcal{D}, \mathcal{Q}) = \sum_{\mathbf{z}^{\,}, \mathbf{z}^+ \in \mathcal{D}} \bar{H}(\mathbf{z}, \mathbf{z^+}, \mathcal{Q}),}
    \label{eq:l_crd}
\end{equation}

\begin{equation}
     \new{\mathcal{L}_{\mathrm{CRD}} = \frac{1}{\lvert \mathcal{D}_s \rvert + \lvert \mathcal{D}_t \rvert} \left[l_{\,_{\mathrm{CRD}}}(\mathcal{D}_s, \mathcal{Q}_t) + l_{\,_{\mathrm{CRD}}}(\mathcal{D}_t, \mathcal{Q}_s) \right].
     \label{eq:L_crd}}
\end{equation}


\begin{figure*}[ht]
\centering
  \includegraphics[width=1\textwidth]{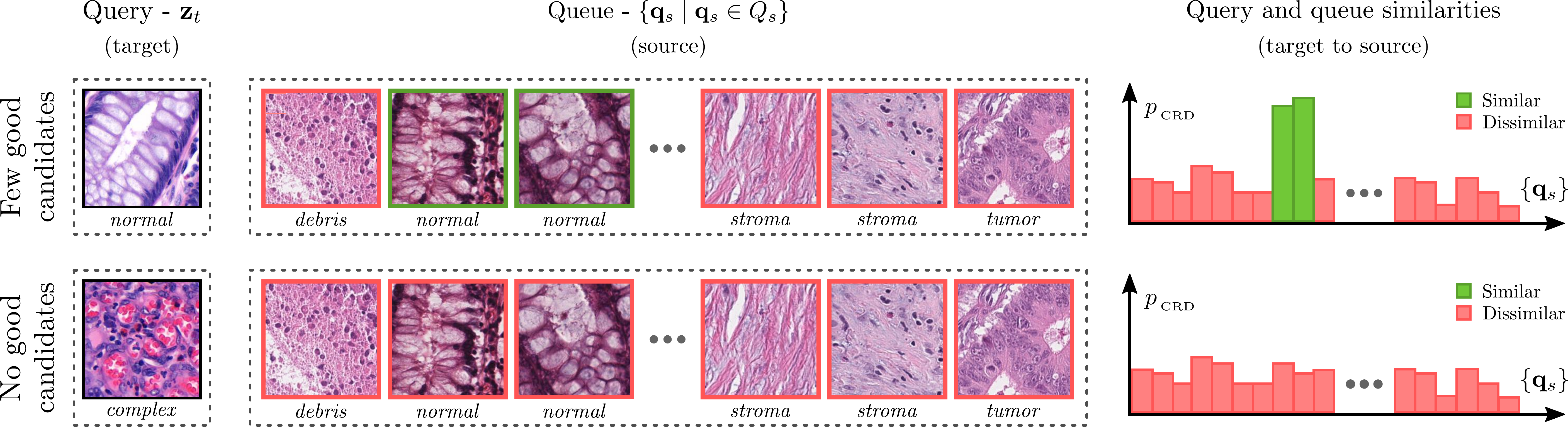}
  \caption{Toy example of the cross-domain matching of different target queries to a fixed source queue. 
  \new{The first column shows two example target query images with computed embedding $\mathbf{z}_t$. 
  The second column depicts the source queue images maintained by the model and their corresponding embeddings $\{\mathbf{q}_s\}$
  In the third column, the distribution of the computed similarities $p_{\,_{\mathrm{CRD}}}$ between the queries and each queue sample are plotted. Similar and dissimilar patterns with respect to the query are displayed in green and red.} 
The top row highlights the case where the model is able to find at least a subset of elements of the queue that match the query (low entropy), as opposed to the bottom row where none of the queue samples match the presented query (high entropy).
  The class labels in this figure have been added for ease of reading and are not available during training.}
  \label{fig:high_low_entropy}
\end{figure*}

\subsection{Easy-to-hard Learning} 
\label{subsec:e2h_learing_scheme}
There are two main pitfalls that can hamper the performance of the cross-domain entropy minimization.

Firstly, at the start of the learning process, the similarity measure between samples and the queue is unclear as the model weights are initialized randomly, which does not guarantee proper feature descriptions. As a result, the optimization of their relative entropy and the loss term $\mathcal{L}_{\mathrm{CRD}}$ is ambiguous in the first epochs.

Secondly, being able to find matching samples for all input queries across datasets is a strong assumption. 
In clinical applications, we often rely on open-source datasets with a limited number of classes to annotate complex tissue databases.
More specifically, challenging tissue types such as complex stroma \new{subtypes} are often not present in public datasets while being frequent in the \acp{WSI} encountered in daily diagnostics. This example is illustrated in Figure~\ref{fig:high_low_entropy}.
The top row shows the case where for a given target query $\mathbf{z_t}$ there are samples with a similar pattern in the source queue, i.e., the distribution of similarities $p_{\,_{\mathrm{CRD}}}$ has low entropy.
The second row highlights the opposite scenario where no queue elements match the query, generating a quasi-uniform distribution of similarities and, therefore, a high entropy. 
In other words, optimizing Equation~\ref{eq:h_mean_crd} for all samples will result in a performance drop as the loss will try to find cross-domain candidates even if there are none to be found.

To tackle both of these issues, we introduce an \acf{E2H} learning scheme.
The model starts with easy to match samples (low \new{cross-entropy}) samples and progressively includes harder (high \new{cross-entropy}) samples as the training progresses.
We assume that the model becomes more robust after each iteration and is more likely to properly process harder examples in later stages. 
Formally, we substitute the domains $\mathcal{D}_{s}, \mathcal{D}_{t}$ in Equation~\ref{eq:L_crd} with the corresponding set of candidates $\mathcal{R}_{s}, \mathcal{R}_{t}$ defined as:

\begin{equation}
    r = \Big\lfloor \frac{e}{N_\mathrm{epochs} \cdot s_{w}} \Big\rfloor \cdot s_{h},
    \label{eq:r}
\end{equation}

\begin{equation} \label{eq:r_set}
\begin{split}
\mathcal{R}_s & = \{\mathbf{z}_{s}^{}, \mathbf{z}_{s}^{+} \in \mathcal{D}_{s} \mid  \bar{H}(\mathbf{z}_{s}^{\,}, \mathbf{z}_{s}^{+}, \mathcal{Q}_{t}) \text{ is reverse top-$r$} \}, \\
\mathcal{R}_t & = \{\mathbf{z}_{t}^{}, \mathbf{z}_{t}^{+} \in \mathcal{D}_{t} \mid  \bar{H}(\mathbf{z}_{t}^{\,}, \mathbf{z}_{t}^{+}, \mathcal{Q}_{s}) \text{ is reverse top-$r$} \},
\end{split}
\end{equation}
            
where the ratio $r$ is gradually increased during training using a step function. We denote $s_{w}$, $s_{h}$ as the width and height of the step, respectively, $N_\mathrm{epochs}$ as the total number of epochs, and 
$e$ the current epoch. \new{The term reverse top-r indicates the ranking of cross-entropy terms in reverse order (low to high values). For example, $r=0.2$ will capture the top $20\%$ of the samples with the lowest cross-entropy.}
This definition ensures that as long as $r=0$ (i.e. $e < N_\mathrm{epochs} \cdot s_{w}$) we only use the in-domain loss $\mathcal{L}_{\mathrm{IND}}$ for backpropagation, and the cross-domain loss term $\mathcal{L}_{\mathrm{CRD}}$ is not considered.
This allows us to first only learn feature representations based on the in-domain feature distribution.
Moreover, with the tuning of the parameter $s_h$ we can \new{control the range of $r$ and thus ensure that its value never reaches $r=1$ to avoid systematic cross-domain matching where no candidates are available.
}

\begin{figure*}[ht]
\centering
  \includegraphics[width=0.9\textwidth]{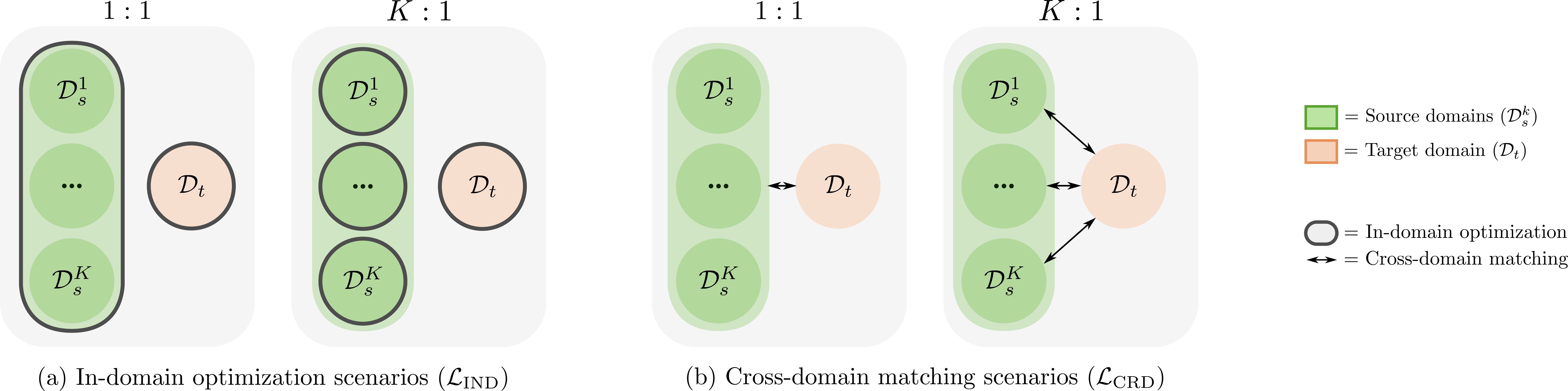}
  \caption{\new{Proposed multi-source scenarios for the in-domain $\mathcal{L}_{\mathrm{IND}}$ (a) and cross-domain $\mathcal{L}_{\mathrm{CRD}}$ (b) optimization. With the one-to-one settings ($1:1$), we treat all source sets $\mathcal{D}_s^k$ as a single set $\mathcal{D}_s$. In the K-to-one ($K:1$) setting, each source domain is considered as an independent set.
  Note that there are no restrictions regarding the combination of the loss terms. For example, the source set can be considered as a single set for the in-domain optimization while being considered as multiple sets for the cross-domain matching.}}
  \label{fig:multi_source_overview}
\end{figure*}

\subsection{Generalization to Multiple Source Scenario}
\label{subsec:generalization_multisource}

Our proposed \ac{SRMA} framework can be generalized to consider multiple datasets in the source domain. 
This is especially useful if the available source datasets overlap in terms of class definitions, which increases the diversity of the visual appearance in the source data.
More formally, we rely on $K$ source datasets denoted $\mathcal{D}_s^k$ where $\bigcup_{k=1}^{K} \mathcal{D}_s^k = \mathcal{D}_s$, and $\mathcal{D} = \mathcal{D}_s \cup \mathcal{D}_t$. The same is valid for the source queues $\mathcal{Q}_s^k$ where $\bigcup_{k=1}^{K} \mathcal{Q}_s^k = \mathcal{Q}_s$, and $\mathcal{Q} = \mathcal{Q}_s \cup \mathcal{Q}_t$. \new{For both the in-domain and cross-domain loss we present two multi-source scenarios as depicted in Figure~\ref{fig:multi_source_overview}.}


\new{One option is to consider the whole source domain as a single domain $\mathcal{D}_s = \bigcup_{k=1}^{K} \mathcal{D}_s^k$ for the in-domain loss:}

\begin{equation}
    \new{\mathcal{L}_{\mathrm{IND}}^{1:1} = \frac{-1}{\lvert \mathcal{D}_s \rvert
    +\lvert \mathcal{D}_t \rvert} \left[ l_{\,_{\mathrm{IND}}}(\bigcup_{k=1}^{K} \mathcal{D}_s^k, \bigcup_{k=1}^{K} \mathcal{Q}_s^k) + l_{\,_{\mathrm{IND}}}(\mathcal{D}_t, \mathcal{Q}_t) \right]}.
    \label{eq:L_ind_11}
\end{equation}

\new{Here, we make no distinction between the source sets and consider a one-to-one features representation importance ($1:1$) between the source and target domain. This definition is equivalent to the single source in-domain adaptation.

Alternatively, we can consider each source and the target domain as independent sets as in Equation~\ref{eq:L_ind_K1}. With this K-to-one ($K:1$) scenario, we have $K+1$ separate in-domain optimizations:}

\begin{equation}
    \new{\mathcal{L}_{\mathrm{IND}}^{K:1} = \frac{-1}{\lvert \mathcal{D}_s \rvert
    +\lvert \mathcal{D}_t \rvert} \left[ \sum_{k=1}^{K} l_{\,_{\mathrm{IND}}}(\mathcal{D}_s^k, \mathcal{Q}_s) + l_{\,_{\mathrm{IND}}}(\mathcal{D}_t, \mathcal{Q}_t) \right]}.
    \label{eq:L_ind_K1}
\end{equation}


\new{The same logic applies to the cross-domain matching. We can either consider a one-to-one correspondence between the unified source domain and the target domain as in Equation~\ref{eq:L_crd_11}, or match each of the individual source domains to the target as in Equation~\ref{eq:L_crd_K1}.}

\begin{equation}
     \new{\mathcal{L}_{\mathrm{CRD}}^{1:1} = \frac{-1}{\lvert \mathcal{D}_s \rvert
    +\lvert \mathcal{D}_t \rvert} \left[l_{\,_{\mathrm{CRD}}}(\bigcup_{k=1}^{K}\mathcal{D}_s^k, \mathcal{Q}_t) + l_{\,_{\mathrm{CRD}}}(\mathcal{D}_t, \bigcup_{k=1}^{K} \mathcal{Q}_s^k) \right].
     \label{eq:L_crd_11}}
\end{equation}

\begin{equation}
     \new{\mathcal{L}_{\mathrm{CRD}}^{K:1} = \frac{-\frac{1}{K}}{\lvert \mathcal{D}_s \rvert
    +\lvert \mathcal{D}_t \rvert} \sum_{k=0}^{K-1}\left[ l_{\,_{\mathrm{CRD}}}(\mathcal{D}_s^k, \mathcal{Q}_t) + l_{\,_{\mathrm{CRD}}}(\mathcal{D}_t, \mathcal{Q}_s^k) \right]}.
     \label{eq:L_crd_K1}
\end{equation}



\new{The formulation of the \ac{E2H} learning procedure has to be updated to comply with multi-source domain definition. For the one-to-one setting, sets $\mathcal{R}_{s}$, $\mathcal{R}_{t}$ remain unchanged as we make no distinction between the different source sets. However, for the K-to-one setting,} the model seeks to match the target domain to the source domain without taking into consideration that there are multiple available source domains. 
We replace the domains $\mathcal{D}_{s}^k, \mathcal{D}_{t}^{}$ in Equation~\ref{eq:L_crd_K1} with the corresponding set of candidates $\mathcal{R}_{s}^{k}, \mathcal{R}_{t}$ defined as:

\begin{equation} \label{eq:r_set_multiple} 
\begin{split}
\mathcal{R}_s^k= \{\mathbf{z}_{s}^{}, \mathbf{z}_{s}^{+} \in \mathcal{D}_{s}^{k} \mid  \bar{H}(\mathbf{z}_{s}^{\,}, \mathbf{z}_{s}^{+}, \mathcal{Q}_{t}) \text{ is reverse top-$r$} \}, \\
\mathcal{R}_t = \{\mathbf{z}_{t}^{}, \mathbf{z}_{t}^{+} \in \mathcal{D}_{t} \mid  \bar{H}(\mathbf{z}_{t}^{\,}, \mathbf{z}_{t}^{+}, \mathcal{Q}_{s}^k) \text{ is reverse top-$r$} \}.
\end{split}
\end{equation}

\new{The overall loss $\mathcal{L}_{\textrm{SRMA}}$ for the multi-source setting is the combination of the in-domain loss ($\mathcal{L}_{\mathrm{IND}}^{1:1}$ or $\mathcal{L}_{\mathrm{IND}}^{K:1}$) and the cross-domain loss ($\mathcal{L}_{\mathrm{CRD}}^{1:1}$ or $\mathcal{L}_{\mathrm{CRD}}^{K:1}$).}

\begin{figure*}[t]
\centering
  \includegraphics[width=.95\textwidth]{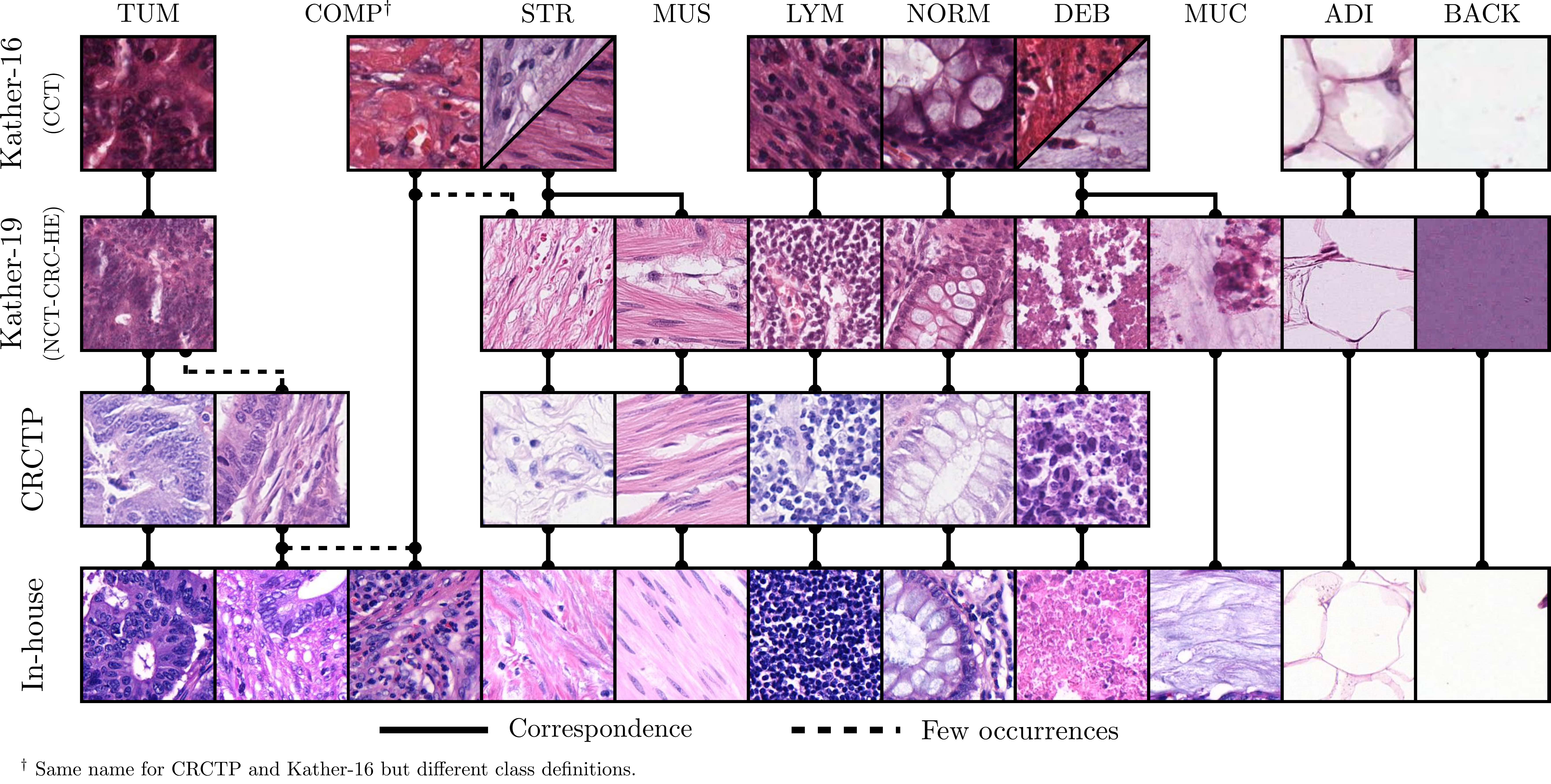}
  \caption{\new{Example images of the different tissue types present in the used datasets and their association. The labeled datasets \acf{K16}, \acf{K19}, and \acf{CRCTP} are publicly available. Examples from the in-house dataset are manually picked for comparison but are not labeled. We use the following abbreviations: TUM: tumor epithelium, STR: simple stroma, COMP: complex stroma, LYM: lymphocytes, NORM: normal mucosal glands, DEB: debris/necrosis, MUS: muscle, MUC: mucus, ADI: adipose tissue, BACK: background. The solid and dashed lines indicate classes correspondences and reported overlaps (also see Section~\ref{subsec:dataset_discr}).}}
  \label{fig:examples-datasets}
\end{figure*}

\section{Datasets}
\label{sec:datasets}

In this study, we use three publicly available datasets, \acf{K16}, \acf{K19} and \acf{CRCTP}, that contain patches extracted from \ac{HE}-stained \acp{WSI} of different tissue types found in the human gastrointestinal tract.
We also use an in-house \ac{CRC} cohort, which does not have patch-level labels, and evaluate our method on three \ac{ROIs}. More details on the datasets can be found below.

Figure~\ref{fig:examples-datasets} shows the occurrence and relationship of different tissue types across all four datasets. The displayed crops of the in-house \ac{WSI} datasets are cherry-picked for comparison purposes.

\subsection{\acl{K16} Dataset}
The \ac{K16} dataset~\citep{kather2016multi} contains $5,000$ patches ($150\times150$ pixels, $74\mu m\times74\mu m$) from multiple \ac{HE} \acp{WSI}. All images are digitized using a scanner magnification of 20x ($0.495\mu m$ per pixel). 
There are eight classes of tissue phenotypes, namely tumor epithelium, simple stroma (homogeneous composition, and smooth muscle), complex stroma (stroma containing single tumor cells and/or few immune cells), immune cells, debris (including necrosis, erythrocytes, and mucus), normal mucosal glands, adipose tissue, and background (no tissue).
The dataset is balanced with 625 patches per class.

\subsection{\acl{K19} Dataset}
The \ac{K19} dataset~\citep{kather2019predicting} consists of patches depicting nine different tissue types: cancerous tissue, stroma, normal colon mucosa, adipose tissue, lymphocytes, mucus, smooth muscle, debris, and background. 
Each class is roughly equally represented in the dataset. In total, there are $100,000$ patches ($224\times224$ pixels, $112\mu m\times112\mu m$) in the training set. All images are digitized using a scanner at a magnification of 20x ($0.5\mu m$ per pixel).

\subsection{\acl{CRCTP} Dataset}
The \ac{CRCTP}~\citep{javed2020cellular} dataset contains a total of $196,000$ patches depicting seven different tissue phenotypes (tumor, inflammatory, stroma, complex stroma, necrotic, benign, and smooth muscle). 
The different phenotypes are roughly equally represented in the dataset. For tumor, complex stroma, stroma, and smooth muscle, there are 35,000 patches per class, for benign and inflammatory, there are 21,000, and for debris, there are 14,000.
The patches ($150\times150$ pixels) are extracted at $20$x resolution from 20 \ac{HE} \acp{WSI}, each one coming from a different patient. For each class, only a subset of the \acp{WSI} is used to extract the patches.  
The annotations are made by two expert pathologists. Out of the two dataset splits available, we use the training set of the patient-level split.

\subsection{In-house Dataset}
Our cohort is composed of 665 \ac{HE}-stained \acp{WSI} from our local \ac{CRC} patient cohort at the Institute of Pathology, University of Bern, Switzerland. The slides originate from 378 unique patients diagnosed with adenocarcinoma and are scanned at a resolution of $0.248 \mu m$ per pixel (40x). None of the selected slides originated from patients that underwent preoperative treatment.

From each \ac{WSI} we uniformly sample 300 ($448\times448$ pixels, $ 111 \mu m \times 111 \mu m$) regions from the foreground masks to reduce the computational complexity of the proposed approach.
This creates a dataset with a total of $199,500$ unique and unlabeled patches. We assume that these randomly selected samples are a good estimation of the tissue complexity and heterogeneity of our cohort. 

We also select three \ac{ROIs} of size $5 \times 5 mm$ ($\simeq 20,000 \times 20,000$ pixel), which are annotated by an expert pathologist according to the definitions used in the \ac{K19} dataset, and use them for evaluation. The regions are selected such that, overall, they represent all tissue types, as well as challenging cases such as late cancer stage (ROI 1), mucinous carcinoma (ROI 2), and torn tissue (ROI 3).

\subsection{Discrepancies in Class Definitions Between Datasets}
\label{subsec:dataset_discr}
The class definitions are not homogeneous across the datasets and they also do not contain the same number of tissue classes.
Following a discussion with expert pathologists, we group stroma/muscle and debris/mucus as stroma and debris, respectively, to create a corresponding adaptation between \ac{K19} and \ac{K16}.

\new{Moreover, the complex stroma class definition between \ac{K16} and \ac{CRCTP} is not identical.} The \ac{CRCTP} complex stroma class contains tiles from the tumor border region and is more consistent with the tumor class in the \ac{K16} and \ac{K19} dataset. In \ac{K16}, the complex stroma is not limited to the tumor border surroundings and is defined as the desmoplastic reaction area, which is usually composed of a mixture of debris, lymphocytes, single tumor cells, and tumor cell clusters.

As a result, the complex stroma class is kept for training but excluded from the evaluation process when performing adaptation on \ac{K16} \new{and \ac{CRCTP}}. 
With this problem definition, we fall into an open-set scenario where the class distribution of the two domains does not rigorously match, as opposed to a closed set adaptation scheme.

\section{Results and Discussion}
\label{sec:results}

In this section, we present and discuss the experimental results. The general experimental setup is described in Section \ref{sec:experimental-setup}.
We validate our proposed self-supervised domain adaptation approach using publicly available datasets and compare it to current \ac{SOTA} methods for \ac{UDA} in Section~\ref{subsec:crossdomain_cls}. 
Additionally, we assess the performance in a clinically relevant use case by validating our model on \ac{WSI} sections from our in-house cohort in Section~\ref{subsec:crossdomain_seg}. 
\new{
We perform an ablation study in Section~\ref{subsec:ablation} for the single-source setting as well as additional experiments on the importance of the \ac{E2H} learning procedure in Section~\ref{subsec:e2h}. 
These experiments are further extended to a multi-source application in Section~\ref{subsec:multi_source_patch}-\ref{subsec:multi_source_wsi} on both publicly available datasets and \ac{WSI} sections.}
To help future research, the implementation and trained models are available open-source\footnote{Code available on \texttt{\href{https://github.com/christianabbet/SRA}{https://github.com/christianabbet/SRA}}.}.

\subsection{General Experimental Setup}
\label{sec:experimental-setup}

In this section, we present the general setup that is used in all experiments.
First, the architecture is trained in an unsupervised fashion, and in a second step, a linear classifier is trained on top as described by \cite{chen2020simple}. 

For the unsupervised learning step, the architecture of the feature extractors, $f_{\Phi}$ and $f_{\Psi}$, are composed of a ResNet18 \citep{he2016deep} followed by two fully connected layers (projection head) using \acp{ReLU}. The output dimension of the multi-layer projection head is $d=128$.
We update the weights of $f_{\Phi}$ as $\theta_{\Phi}$ using standard backpropagation and $f_{\Psi}$ as $\theta_{\Psi}$ with momentum $m=0.999$, as described in \cite{he2020momentum}.


The model is trained from scratch for $N_\mathrm{epochs}=200$ epochs until convergence using the \ac{SGD} optimizer ($\text{momentum}=0.9$, $\text{weight decay}=10^{-4}$), a learning rate of $\lambda=0.03$, and a batch size of $B=128$. The size of the queue is fixed to $\vert \mathcal{Q} \rvert = 2^{16} = 65,536$ samples.
For the similarity learning we set $\tau=0.2$.
We apply \new{random cropping, grayscale transformation, horizontal/vertical flipping, rotation, grid distortion, ISO noise, Gaussian noise, and color jittering} as data augmentations $f_T$. 
At each epoch, we sample $50,000$ examples with replacement from both the source and target dataset to create $\mathcal{D}$ with a total of $N=100,000$ samples. \new{The ratio $r$ is updated between each epoch, while the sets $\mathcal{R}_s$, $\mathcal{R}_t$ for cross-domain matching are computed batch-wise}.

During the second phase, the momentum encoder branch is discarded as it is not used for inference.
The classification performance is evaluated using a linear classifier, which is placed on top of the frozen feature extractor. The linear classifier directly matches the output of the embedding $d$ to the number of classes. It is trained for $N_{\textrm{epochs}}=100$ epochs until convergence using the \ac{SGD} optimizer ($\text{momentum}=0.9$, $\text{weight decay}=0$), a batch size of $B=128$, and a learning rate of $\lambda=1$. We use only few randomly selected source labels to train this classification layer in order to simulate the clinical application, where we usually rely on a large quantity of unlabeled data and only have access to few labeled samples. \new{More precisely, we use $n=1,000$ samples (i.e., $1\%$) to train the linear classifier with \ac{K19} and $n=500$ samples (i.e., $10\%$) when training with \ac{K16}.} While training the linear classifier, we multi-run $10$ times to obtain statistically relevant results.
The set of selected source labels varies between these runs, as they are randomly sampled for each run.
If not specified otherwise, we use $s_w=0.25$ and $s_h=0.15$ for \ac{E2H} learning. 

For a fair comparison, we also use a ResNet18 backbone for all the presented baselines.

\subsection{Cross-Domain Patch Classification}
\label{subsec:crossdomain_cls}

In this task, we use the larger dataset \ac{K19} as the source dataset and adapt it to \ac{K16}. We motivate the selection of \ac{K19} as the source set by the fact that it is closer to the clinical scenario where we mainly rely on a large quantity of unlabeled data and only a few labeled ones, by using only $1\%$ of the labels in \ac{K19}. We evaluate the performance of the model with the patch classification task on the \ac{K16} dataset. The mucin and muscle in \ac{K19} are grouped with debris and stroma, respectively, to allow comparison with the \ac{K16} class definitions. We use $70\%$ of \ac{K16} to train the unsupervised domain adaptation. The remaining $30\%$ are used to test the performance of the linear classifier trained on top of the self-supervised model. 

\begin{figure*}[ht!]
\centering
  \includegraphics[width=0.99\textwidth]{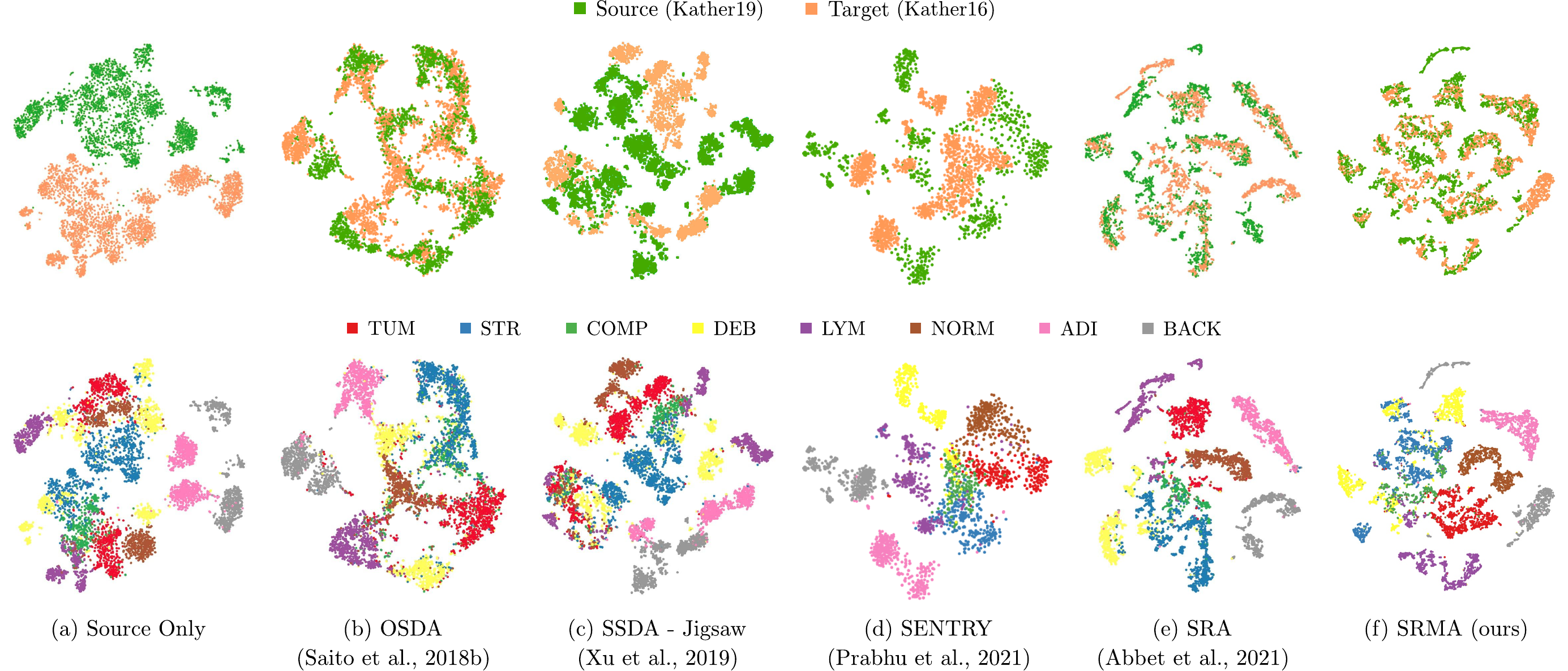}
  \caption{\new{The t-SNE projection of the source (\ac{K19}) and target (\ac{K16}) domain embeddings. The top row shows the alignment between the source and target domain, while the bottom row highlights the representations of the different classes. We compare our approach (f) to other \ac{UDA} methods (b-e), and the fully supervised, transfer learning baseline (source only) (a).}}
  \label{fig:tsne}
\end{figure*}

The results of our proposed \ac{SRMA} method are presented in Table~\ref{tbl:cross-domain}, in comparison with baselines and \ac{SOTA} algorithms for domain adaption. 
As the lower bound, we consider \new{both MoCoV2 where the source and the target domain are merged into a single set} and direct transfer learning (source only), where the model is trained in a supervised fashion on the source data only. 
We use the same logic for the upper bound by training on the target domain data (fully supervised approach). The performances on complex stroma are not reported as the class is not present in \ac{K19}. Figure~\ref{fig:tsne} shows the t-SNE projection and alignment of the domain adaptation for the transfer learning (source only), the top-performing baselines (OSDA, SSDA with jigsaw solving), and our method (\ac{SRMA}). 
Complementary results can be found in \ref{app:selfsupmodel} and \ref{app:tsne}. 

\begin{table*}[ht!]
    \caption{Results of the domain adaptation from \ac{K19} (source) to \ac{K16} (target). 1\% of the source domain labels are \new{used} and the target domain labels are unknown. Complex stroma is excluded as the class is not present in \ac{K19}. \new{The mucin and muscle class in \ac{K19} are grouped with debris and stroma, respectively, as they overlap in \ac{K16}}. The top results for the domain adaptation methods are highlighted in bold. 
    We report the F1 score for each class as well as the overall weighted F1 score averaged over 10 runs.}
    \label{tbl:cross-domain}
    \centering
    \begin{footnotesize}
    \begin{threeparttable}
    \begin{tabular}{l l c c c c c c c c c}
    \toprule
    \toprule
    Methods & TUM & COMP & STR & LYM & DEB & NORM & ADI & BACK & ALL\\
    \midrule
    \new{MoCoV2 \citep{chen2020improved}\tnote{$\dagger$}} &
    \new{36.8\tnote{**}} & \new{-} & \new{45.4\tnote{**}} & \new{27.1\tnote{**}} & \new{30.8\tnote{**}} & \new{45.2\tnote{**}} & \new{43.1\tnote{**}} & \new{43.6\tnote{**}} & \new{38.9\tnote{**}} \\
    Source only$^{\ddagger}$& 74.0\tnote{**} & - & 77.4\tnote{**} & 75.3\tnote{**} & 50.5\tnote{**} & 66.9\tnote{**} & 87.0\tnote{**} & 93.1\tnote{**} & 75.1\tnote{**} \\
    \midrule
    DANN \citep{ganin2015unsupervised} & 65.8\tnote{**} & - & 60.8\tnote{**} & 42.3\tnote{**} & 47.8\tnote{**} & 61.9\tnote{**} & 64.1\tnote{**} & 62.3\tnote{**} & 57.8\tnote{**}\\
    Stain norm. \citep{macenko2009method} & 77.8\tnote{**} & - & 75.9\tnote{**} & 68.2\tnote{**} & 42.1\tnote{**} & 75.1\tnote{**} & 77.4\tnote{**} & 87.6\tnote{**} & 72.2\tnote{**} \\
    CylceGAN \citep{zhu2017unpaired} & 70.7\tnote{**} & - & 71.6\tnote{**} & 62.3\tnote{**} & 47.6\tnote{**} & 75.5\tnote{**} & 89.0\tnote{**} & 88.2\tnote{**} & 72.4\tnote{**} \\
    SelfPath \citep{koohbanani2020self}& 71.5\tnote{**} & - & 68.8\tnote{**} & 68.1\tnote{**} & 57.6\tnote{**} & 77.6\tnote{**} & 82.3\tnote{**} & 85.5\tnote{**} & 73.1\tnote{**} \\
    OSDA \citep{saito2018open} & 82.0\tnote{**} & - & 78.2\tnote{*} & 83.6\tnote{*} & 63.8\tnote{**} & 80.3\tnote{**} & 90.8\tnote{**} & 93.2\tnote{*} & 81.7\tnote{**}  \\
    SSDA - Rot \citep{xu2019self} & 85.1\tnote{**} & - & 78.5\tnote{**} & 81.3\tnote{**} & \textbf{68.2} & 88.7\tnote{**} & 93.9\tnote{**} & 96.5\tnote{**} & 84.7\tnote{**}  \\
    SSDA - Jigsaw \cite{xu2019self} & 90.0\tnote{**} & - & \textbf{81.2} & 79.5\tnote{**} & 64.4\tnote{**} & 88.3\tnote{**} & 94.2\tnote{**} & 95.7\tnote{*} & 84.9\tnote{**}  \\
    \new{SENTRY \citep{prabhu2021sentry}} & \new{88.7\tnote{**}} & \new{-} & \new{74.4\tnote{**}} & \new{\textbf{86.0}} & \new{\textbf{65.5}\tnote{+}} & \new{91.5\tnote{**}} & \new{94.1\tnote{**}} & \new{\textbf{97.9}\tnote{+}} & \new{85.7\tnote{**}}  \\
    SRA \citep{abbet2021selfrule} & 93.4\tnote{**} & - & 72.9\tnote{**} & 82.7\tnote{*} & \textbf{67.9}\tnote{+} & 96.5\tnote{*} & \textbf{97.0}\tnote{+} & \textbf{97.2}\tnote{+} & 86.9\tnote{*} \\
    \new{SRMA (ours)} & \new{\textbf{97.3}} & \new{-} & \new{\textbf{79.3}\tnote{+}} & \new{80.2\tnote{**}} & \new{62.2\tnote{**}} & \new{\textbf{98.7}} & \new{\textbf{97.6}} & \new{\textbf{98.1}} & \new{\textbf{87.7}} \\
    \midrule
    Target only$^{\mathsection}$ & 94.6\tnote{**} & - & 83.6\tnote{**} & 92.6\tnote{**} & 88.7\tnote{**} & 95.4\tnote{**} & 97.8\tnote{+} & 98.5\tnote{+} & 93.0\tnote{**} \\
    \bottomrule
    \bottomrule
    \end{tabular}
    \begin{tablenotes}
        \item[$\dagger$] Source and target dataset are merged and trained using contrastive learning.
        \item[$\ddagger$] Direct transfer learning: trained on the source domain only, no adaptation (lower bound).
        \item[$\mathsection$] Fully supervised: trained knowing the labels of the target domain (upper bound).
        \item[+] $\ p\geq0.05$; $^{*}\ p<0.05$; $^{**}\ p<0.001$; unpaired t-test with respect to the top result.
    \end{tablenotes}
    \end{threeparttable}
    \end{footnotesize}
\end{table*}

\new{MoCoV2 fails to generalize knowledge between source and target domain. Since the model is not constrained, it learns two distinct embeddings for each domain. The experiment highlights the limitations of contrastive learning without domain adaptation.}

Stain normalization slightly decreases the performances, compared to the source only baseline, as it introduces color artifacts that are very challenging for the network classifier. 
This mainly comes from the distribution of target samples, namely \ac{K16}, that are composed of dark stained patches which are difficult to normalize properly.

CycleGAN suffers from performance degradation for the lymphocytes predictions. Like color normalization, it tends to create saturated images. In addition, the model alters the shape of the lymphocytes nuclei, thus fooling the classifier toward either debris or tumor classification. 

In our setup, we observe that the use of the gradient reversal layer leads to an unstable loss optimization for both Self-Path and DANN, which explains the large performance drops when training. Heavier data augmentations partially solve this issue.

OSDA benefits from the open-set definition of the approach and achieves very good performance for lymphocytes detections.

\begin{figure*}[!htb]
\centering
  \includegraphics[width=1.0\textwidth]{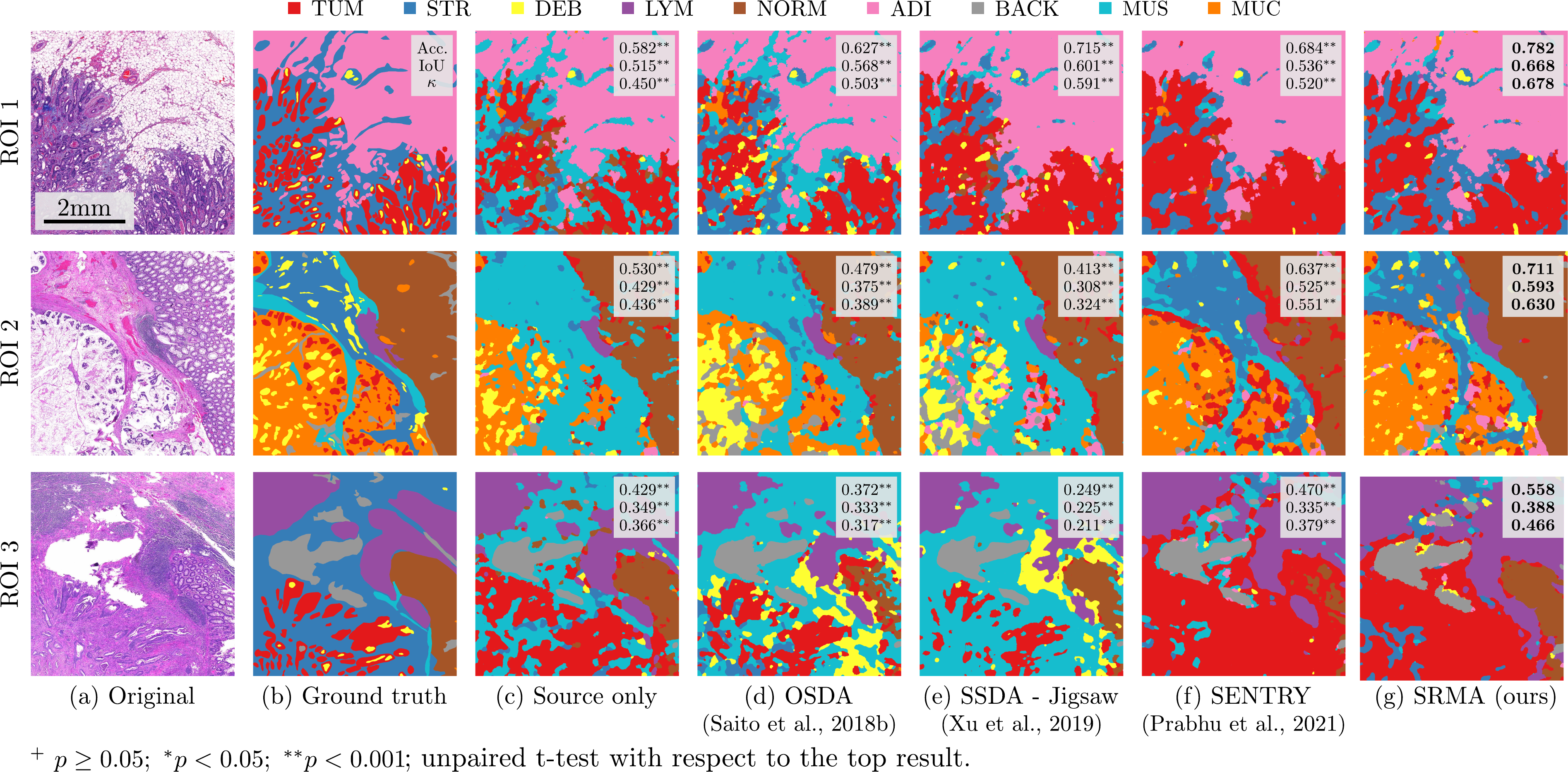}
  \caption{\new{Quantitative results of the domain adaptation from \ac{K19} to our unlabeled in-house dataset based on three selected \acf{ROIs}.} (a-b) show the original \ac{ROIs} from the \acp{WSI} and their ground truth, respectively. We compare the performance of our \acf{SRMA} algorithm (g) to the lower bound and the top-performing \ac{SOTA} methods (c-f). We report the pixel-wise accuracy, the weighted \acl{IoU}, and the pixel-wise Cohen's kappa ($\kappa$) score averaged over 10 runs.}
  \label{fig:bern_wsi}
\end{figure*}

SSDA achieves similar results when using either rotation or jigsaw puzzle-solving as an auxiliary task. Due to the rotational invariance structure of the tissue and selected large magnification for tilling, rotation and jigsaw puzzle-solving are not optimal auxiliary tasks for digital pathology.

\new{Out of the presented baselines, SENTRY achieves top competitive results on almost all classes. The main limitation appears to be the distinction between tumor and normal mucosa.}

Our proposed \ac{SRMA} method shows an excellent alignment between the same class clusters of the source and target distributions and outperforms \ac{SOTA} approaches in terms of weighted F1 score. 
Notably, our approach is even able to match the upper bound model for normal and tumor tissue identification.
The embedding of complex stroma, which only exists in the target domain, is represented as a single cluster with no matching candidates, which highlights the model's ability to reject unmatchable samples from domain alignment.

Furthermore, the cluster representation is more compact
compared to other presented methods, where for example, normal mucosa tends to be aligned with complex stroma and tumor.
Our approach suffers a drop in performance for stroma detection, which can be explained by the presence of lymphocytes in numerous stroma tissue examples, causing a higher rate of misclassification. Moreover, the presence of loose tissue that has a similar structure as stroma in the debris class is challenging. The overlap is also observed in the embedding projection.

\subsection{Use Case: Cross-Domain Segmentation of WSIs}
\label{subsec:crossdomain_seg}

To further validate our approach in a real case scenario, we perform domain adaptation using our proposed model from \ac{K19} to our in-house dataset and validate it on \acp{WSI} \acf{ROIs}. 

The results are presented in Figure~\ref{fig:bern_wsi}, alongside the original \ac{HE} \ac{ROIs}, their corresponding ground truth annotations, direct transfer learning (source only), 
as well as comparative results of the top-scoring \ac{SOTA} approaches. We use a tile-based approach to predict classes on each \ac{ROIs}
and use conditional random fields as in \cite{chan2019histosegnet} to smooth the prediction map. 
The number of available labeled tissue regions is limited to the presented \ac{ROIs}.

For all models, stroma and muscle are poorly differentiated as both have similar visual features without contextual information. 
This phenomenon is even more apparent in the source only setting, where muscle tissue is almost systematically interpreted as stroma. Moreover, due to the lack of domain adaptation, the boundary between tumor and normal tissues is not well defined, leading to incorrect predictions of these classes. 

\begin{figure*}[!htb]
\centering
  \includegraphics[width=0.9\textwidth]{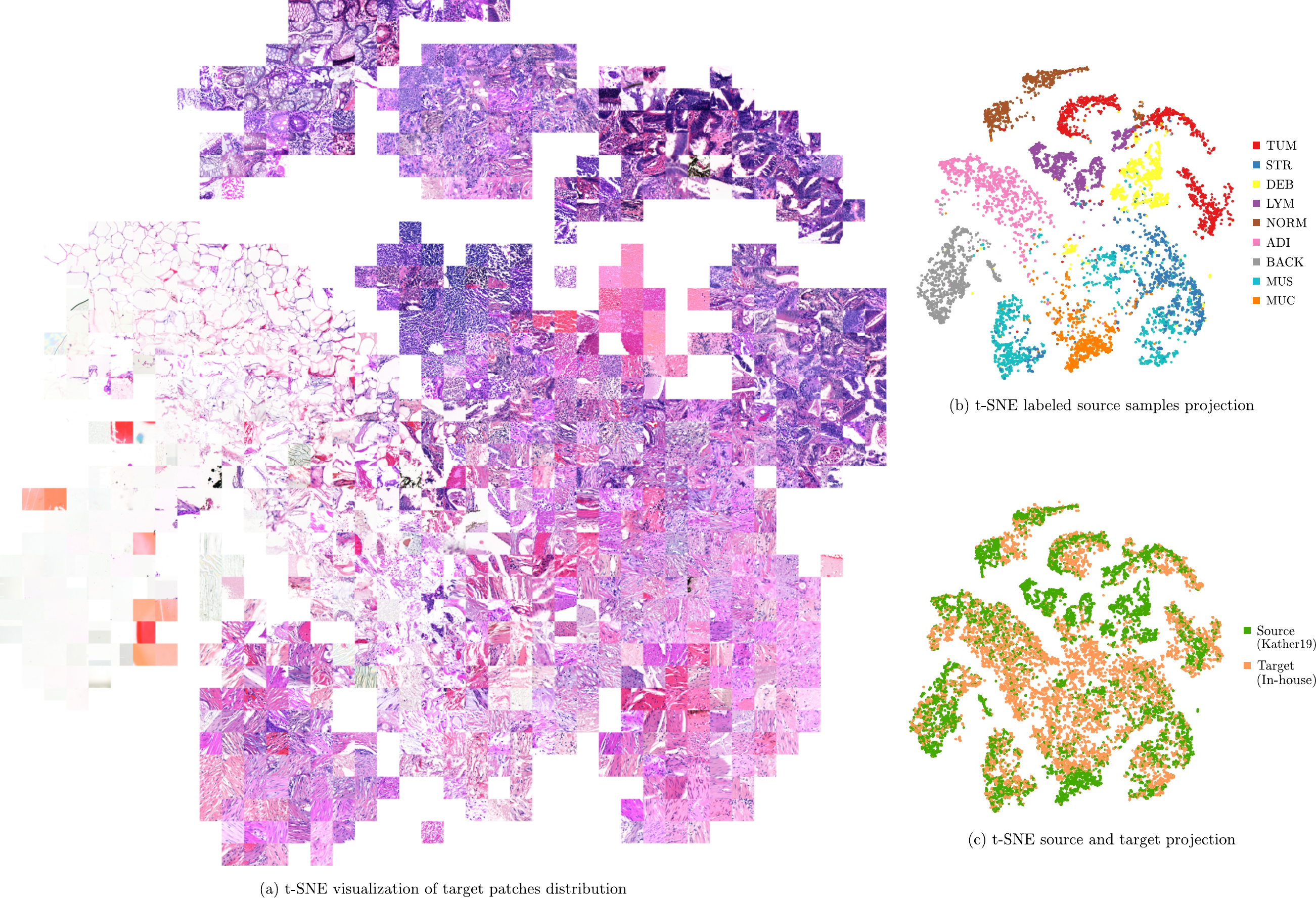}
  \caption{The t-SNE visualization of the \ac{SRMA} model trained on \ac{K19} and our in-house data. All sub-figures depict the same embedding. 
  (a) Patch-based visualization of the embedding.
  (b) Distribution of the labeled source samples.
  (c) The relative alignment of the source and target domain samples.}
  \label{fig:k19_bern_tsne_img}
\end{figure*}

OSDA, on the other hand, fails to adapt and generalize to new tumor examples while trying to reject mistrusted samples. This phenomenon is most visible in ROI 3, where the model interprets the surroundings of the cancerous region as a mixture of debris, stroma, and muscle.  

SSDA tends to predict lymphocyte aggregates as debris. This can be explained by the model's sensitivity to staining variations as well as both classes' similarly dotted structure. Moreover, the model struggles to properly embed the representations of mucin. The scarcity of mucinous examples in the target domain makes the representation of this class difficult.

\new{As in the patch classification task, SENTRY is as the top performing baseline. However, the model is still limited by its capacity to distinguish between tumor and normal mucosa due to the few label setting. Also, the detection of the stroma area appears less detailed compared to other approaches such as OSDA or \ac{SRMA}.
}

Our approach outperforms the other \ac{SOTA} domain adaptation methods in terms of pixel-wise accuracy, weighed \ac{IoU} and pixel-wise Cohen's kappa score $\kappa$.
Regions with mixed tissue types (e.g., lymphocytes + stroma or stroma + isolated tumor cells) are challenging cases because the samples available in the public cohorts mainly contain homogeneous tissue textures and few examples of class mixtures. Subsequently, domain adaptation methods naturally struggle to align features resulting in a biased classification. 
We observe that thinner or torn stroma regions, where the background behind is well visible, are often misclassified as adipose tissue by \ac{SRMA}, which is most likely due to their similar appearance. However, our \ac{SRMA} model is able to correctly distinguish between normal mucosa and tumor, which are tissue regions with very relevant information downstream tasks such as survival analysis.

Figure~\ref{fig:k19_bern_tsne_img} presents a qualitative visualization of the model's embedding space. The figure shows the actual visual distribution of the target patches, the source domain label arrangement, and the overlap of the source and target domain. The patch visualization also shows a smooth transition between class representations where for example, neighboring samples of the debris cluster include a mixture of tissue and debris. The embedding reveals a large area in the center of the visualization that does not match with the source domain. The area mostly includes loose connective tissue and stroma, which are both under-represented in the training examples. Also, mucin is improperly matched to the loose stroma, which explains the misclassification of stromal tissue in the ROI 2. The scarcity of mucinous examples in our in-house cohort makes it difficult for the model to find good candidates.

\begin{table*}[htb]
    \caption{Ablation study for the proposed \acf{SRMA} approach. We denote $\mathcal{L}_{\mathrm{IND}}$ as the in-domain loss, $\mathcal{L}_{\mathrm{CRD}}$ as the cross-domain loss, and E2H as easy-to-hard. We train the domain adaptation from \acl{K19}  \new{(source) to \acl{K16} (target). Only $1\%$ of the source domain labels are used, and no labels for the target domain. We report the F1 and weighted F1 score for the individual classes and the overall mean performance (all) (average over 10 runs).}}
    \label{tbl:cross-domain-ablation}
    \centering
    \begin{footnotesize}
    \begin{threeparttable}
    \begin{tabular}{lcccccrrrrrrrrrr}
    \toprule
    \toprule
    \new{Methods} & & $\mathcal{L}_{\mathrm{IND}}$ & $\mathcal{L}_{\mathrm{CRD}}$ & E2H & & TUM & STR & LYM & DEB & NORM & ADI & BACK & ALL\\
    \midrule
    \new{MoCoV2\tnote{$\dagger$}}& & - & - & - & & 
    36.8\tnote{**} & 45.4\tnote{**} & 27.1\tnote{**} & 30.8\tnote{**} & 45.2\tnote{**} & 43.1\tnote{**} & 43.6\tnote{**} & 38.9\tnote{**} 
    \\
    \new{SRA\tnote{$\ddagger$}} & & $\checkmark$ & - & - & &
    88.1\tnote{**} & 72.8\tnote{**} & 78.0\tnote{*} & 71.8\tnote{*} & 89.9\tnote{**} & 93.4\tnote{*} & 86.0\tnote{*} & 82.9\tnote{**}
    \\
    \new{SRA\tnote{$\ddagger$}} & & - & $\checkmark$ & - & & 
    14.1\tnote{**} & 9.1\tnote{**} & 0.2\tnote{**} & 10.1\tnote{**} & 4.9\tnote{**} & 0.0\tnote{**} & 61.5\tnote{**} & 14.4\tnote{**} 
    \\
    \new{SRA\tnote{$\ddagger$}} & & $\checkmark$ & $\checkmark$ & - & & 
    63.0\tnote{**} & 69.9\tnote{**} & \textbf{85.1} & 57.7\tnote{**} & \textbf{98.2\tnote{+}} & \textbf{97.9} & 90.0\tnote{**} & 80.3\tnote{**} 
    \\
    \new{SRA\tnote{$\ddagger$}} & & $\checkmark$ & $\checkmark$ & $\checkmark$ & & 
    93.4\tnote{**} & 72.9\tnote{**} & 82.7\tnote{*} & \textbf{67.9} & 96.5\tnote{**} & 97.0\tnote{**} & 97.2\tnote{*} & 86.9\tnote{*}
    \\
    \new{SRMA} & & - & $\checkmark$ & - & & 
    \new{35.3\tnote{**}} & \new{3.6\tnote{**}} & \new{0.0\tnote{**}} & \new{2.1} & \new{15.6\tnote{**}} & \new{64.0\tnote{**}} & \new{16.5\tnote{**}} & \new{19.8\tnote{**}}
    \\
    \new{SRMA} & & $\checkmark$ & $\checkmark$ & - & &
    \new{93.3\tnote{**}} & \new{\textbf{77.4}\tnote{+}} & \new{80.5\tnote{**}} & \new{\textbf{66.2}\tnote{+}} & \new{91.4\tnote{**}} & \new{\textbf{97.8}\tnote{+}} & \new{\textbf{98.3}} & \new{86.5\tnote{*}}
    \\
    \new{SRMA} & & $\checkmark$ & $\checkmark$ & $\checkmark$ & & 
    \new{\textbf{97.3}} & \new{\textbf{79.3}} & \new{80.2\tnote{**}} & \new{62.2\tnote{**}} & \new{\textbf{98.7}} & \new{\textbf{97.6}\tnote{+}} & \new{\textbf{98.1}\tnote{+}} & \new{\textbf{87.7}}
    \\
    \bottomrule
    \bottomrule
    \end{tabular}
    \begin{tablenotes}
        \item[$\dagger$] \new{\cite{chen2020improved}. Source and target dataset are merged and trained using contrastive learning.}
        \item[$\ddagger$] \new{\cite{abbet2021selfrule}.}
        \item[+] $\ p\geq0.05$; $^{*}\ p<0.05$; $^{**}\ p<0.001$; unpaired t-test with respect to top result.
    \end{tablenotes}
    \end{threeparttable}
    \end{footnotesize}
\end{table*}

\begin{table*}[htb]
    \caption{Ablation study for the proposed \acf{SRMA} approach. We denote $\mathcal{L}_{\mathrm{IND}}$ as the in-domain loss, $\mathcal{L}_{\mathrm{CRD}}$ as the cross-domain loss, and E2H as easy-to-hard. We train the domain adaptation from \acl{K19} \new{(source) to our in-house dataset (target). Only $1\%$ of the source domain labels are used, and no labels for the target domain. We report the pixel-wise accuracy, the weighted intersection over union, and the pixel-wise Cohen’s kappa ($\kappa$) score for three manually annotated \acf{ROIs} (average over 10 runs)}.}
    \label{tbl:cross-domain-ablation-wsi}
    \centering
    \scriptsize
    \begin{threeparttable}
    \begin{tabular}{lcccc@{\hskip 0.0in}rrrc@{\hskip 0.0in}rrrc@{\hskip 0.0in}rrrc@{\hskip 0.0in}rrr}
    \toprule
    \toprule
    &&&&& \multicolumn{3}{c}{ROI 1} && \multicolumn{3}{c}{ROI 2} && \multicolumn{3}{c}{ROI 3}&& \multicolumn{3}{c}{ROI 1-3} \\
    \cmidrule{6-8} 
    \cmidrule{10-12} 
    \cmidrule{14-16}
    \cmidrule{18-20}
    \new{Methods} & $\mathcal{L}_{\mathrm{IND}}$ & $\mathcal{L}_{\mathrm{CRD}}$ & E2H && Acc. & IoU & $\kappa$ & & Acc. & IoU & $\kappa$ & & Acc. & IoU & $\kappa$ & & Acc. & IoU & $\kappa$ \\
    \midrule
    \new{MoCoV2 \tnote{$\dagger$}} & \new{-} & \new{-} & \new{-} & &
    \new{0.556\tnote{**}} & 
    \new{0.470\tnote{**}} & 
    \new{0.417\tnote{**}} & &
    \new{0.298\tnote{**}} & 
    \new{0.198\tnote{**}} & 
    \new{0.220\tnote{**}} & &
    \new{0.321\tnote{**}} & 
    \new{0.255\tnote{**}} & 
    \new{0.240\tnote{**}} & & 
    \new{0.399\tnote{**}} & 
    \new{0.301\tnote{**}} & 
    \new{0.319\tnote{**}} 
    \\
    \new{SRA \tnote{$\ddagger$}} & \new{$\checkmark$} & \new{-}& \new{-} & &
    \new{0.754\tnote{*}} & 
    \new{\textbf{0.655}\tnote{+}} & 
    \new{0.646\tnote{*}} & & 
    \new{0.679\tnote{*}} & 
    \new{0.551\tnote{*}} & 
    \new{0.594\tnote{*}} & & 
    \new{0.498\tnote{**}} & 
    \new{0.357\tnote{**}} & 
    \new{0.415\tnote{**}} & & 
    \new{0.644\tnote{**}} & 
    \new{0.497\tnote{**}} & 
    \new{0.590\tnote{**}}  
    \\
    \new{SRA \tnote{$\ddagger$}}  & \new{-} & \new{$\checkmark$} & \new{-} & &
    \new{0.108\tnote{**}} & 
    \new{0.022\tnote{**}} & 
    \new{0.000\tnote{**}} & & 
    \new{0.060\tnote{**}} & 
    \new{0.004\tnote{**}} & 
    \new{0.000\tnote{**}} & & 
    \new{0.061\tnote{**}} & 
    \new{0.006\tnote{**}} & 
    \new{0.000\tnote{**}} & & 
    \new{0.076\tnote{**}} & 
    \new{0.008\tnote{**}} & 
    \new{0.000\tnote{**}}  
    \\
    \new{SRA \tnote{$\ddagger$}} & \new{$\checkmark$} & \new{$\checkmark$} & \new{-} & &
    \new{0.766\tnote{*}} & 
    \new{\textbf{0.660}\tnote{+}} & 
    \new{0.658\tnote{*}} & &
    \new{0.701\tnote{*}} & 
    \new{0.582\tnote{*}} & 
    \new{0.619\tnote{*}} & &
    \new{0.526\tnote{**}} & 
    \new{0.368\tnote{*}} & 
    \new{0.438\tnote{**}} & & 
    \new{0.664\tnote{**}} & 
    \new{0.526\tnote{*}} & 
    \new{0.615\tnote{**}} 
    \\
    \new{SRA \tnote{$\ddagger$}} & \new{$\checkmark$} & \new{$\checkmark$} & \new{$\checkmark$} & &
    \new{0.752\tnote{*}} & 
    \new{0.638\tnote{*}} & 
    \new{0.639\tnote{*}} & &
    \new{0.689\tnote{**}} & 
    \new{0.574\tnote{**}} & 
    \new{0.607\tnote{**}} & &  
    \new{0.541\tnote{*}} & 
    \new{0.373\tnote{*}} & 
    \new{0.448\tnote{*}} & &
    \new{0.661\tnote{**}} & 
    \new{0.521\tnote{**}} & 
    \new{0.611\tnote{**}} 
    \\
    \new{SRMA} & \new{-} & \new{$\checkmark$} & \new{-} & &
    \new{0.593\tnote{**}} & 
    \new{0.471\tnote{**}} & 
    \new{0.429\tnote{**}} & & 
    \new{0.096\tnote{**}} & 
    \new{0.019\tnote{**}} &
    \new{0.029\tnote{**}} & & 
    \new{0.261\tnote{**}} & 
    \new{0.118\tnote{**}} & 
    \new{0.080\tnote{**}} & &
    \new{0.322\tnote{**}} & 
    \new{0.166\tnote{**}} & 
    \new{0.196\tnote{**}} 
    \\
    \new{SRMA} & \new{$\checkmark$} & \new{$\checkmark$} & \new{-} & &
    \new{0.724\tnote{**}} & 
    \new{0.634\tnote{**}} & 
    \new{0.608\tnote{**}} & &
    \new{\textbf{0.706}\tnote{+}} & 
    \new{\textbf{0.591}\tnote{+}} &
    \new{\textbf{0.630}\tnote{+}} & &
    \new{0.518\tnote{**}} & 
    \new{0.319\tnote{**}} & 
    \new{0.415\tnote{**}} & &
    \new{0.650\tnote{**}} & 
    \new{0.484\tnote{**}} & 
    \new{0.599\tnote{**}} 
    \\
    \new{SRMA} & \new{$\checkmark$} & \new{$\checkmark$} & \new{$\checkmark$} & &
    \new{\textbf{0.782}} & 
    \new{\textbf{0.668}} & 
    \new{\textbf{0.678}} & &
    \new{\textbf{0.711}} & 
    \new{\textbf{0.593}} & 
    \new{\textbf{0.630}} & &  
    \textbf{\new{0.558}} & 
    \textbf{\new{0.388}} & 
    \textbf{\new{0.466}}  & & 
    \new{\textbf{0.684}} & 
    \new{\textbf{0.535}} & 
    \new{\textbf{0.635}} 
    \\
    \bottomrule
    \bottomrule
    \end{tabular}
    \begin{tablenotes}
        \item[$\dagger$] \new{\cite{chen2020improved}. Source and target dataset are merged and trained using contrastive learning.}
        \item[$\ddagger$] \new{\cite{abbet2021selfrule}.}
        \item[+] $\ p\geq0.05$; $^{*}\ p<0.05$; $^{**}\ p<0.001$; unpaired t-test with respect to top result.
    \end{tablenotes}
    \end{threeparttable}
\end{table*}

\subsection{Ablation Study of the Proposed Loss Function}
\label{subsec:ablation}

In this section, we present the ablation study of our \ac{SRMA} approach. 
We denote $\mathcal{L}_{\mathrm{IND}}$ as the in-domain loss, $\mathcal{L}_{\mathrm{CRD}}$ as the cross-domain loss, and \ac{E2H} as the easy-to-hard learning scheme.
We evaluate the performance of our model on two tasks. \new{The first one is the domain alignment between \ac{K19} (source) and \ac{K16} (target), which follows the experimental setting described in Section \ref{subsec:crossdomain_cls}. The results are presented in Table~\ref{tbl:cross-domain-ablation}. The second task is the domain adaptation of \ac{K19} (source) to \ac{ROIs} from our in-house dataset (target), as presented in Section \ref{subsec:crossdomain_seg}. Table~\ref{tbl:cross-domain-ablation-wsi} shows the results of these experiments.
The following section jointly discusses the results of both tasks.}

\new{
We use MoCoV2 \citep{chen2020improved} as a baseline, where both domains are merged to a dataset $\mathcal{D}$, and train following a contrastive learning approach. 
We also compare our proposed approach \ac{SRMA} to our previous work \ac{SRA} \citep{abbet2021selfrule}. 
For the single-source domain adaptation, the difference between \ac{SRA} and the proposed extension \ac{SRMA} lies in the reformulation of the cross-entropy matching. As a result, only the entropy-related terms, namely $\mathcal{L}_{\mathrm{CRD}}$ and \ac{E2H}, are affected. Thus, training \ac{SRA} and \ac{SRMA} using only the in-domain loss $\mathcal{L}_{\mathrm{IND}}$ is the same set-up.
}

The baseline fails to learn discriminant features that match both sets leading to poor performances in both cross-domain adaptation tasks. This shows that, if not constrained, the model is not able to generalize the knowledge and ends up learning two distinct feature spaces, one for the source and one for the target domain.

\begin{figure}
  \begin{minipage}[b]{0.7\textwidth}
    \centering
      
    \scriptsize
    \begin{threeparttable}
    \begin{tabular}{lrrrcrrrcrrr}
    \toprule
    \toprule
     & \multicolumn{11}{c}{Metrics}  \\
    \cmidrule{2-12}
     Images & Acc. & IoU & $\kappa$ & & Acc. & IoU & $\kappa$ & & Acc. & IoU & $\kappa$\\
    \midrule
    & \multicolumn{3}{l}{\tiny $s_w=0.125\,, s_h = 0.075$} & & \multicolumn{3}{l}{\tiny $s_w=0.125\,, s_h = 0.1$} & & \multicolumn{3}{l}{\tiny $s_w=0.125\,, s_h = 0.125$}\\
    \cmidrule{2-4}
    \cmidrule{6-8}
    \cmidrule{10-12}
    ROI 1& 
    \new{\textbf{0.777}\tnote{+}} & 
    \new{0.658\tnote{*}} & 
    \new{\textbf{0.670}\tnote{+}} && 
    \new{0.758\tnote{*}} & 
    \new{0.642\tnote{**}} & 
    \new{0.646\tnote{**}} && 
    \new{0.752\tnote{**}} & 
    \new{0.652\tnote{*}} & 
    \new{0.643\tnote{**}} \\
    ROI 2&  
    \new{0.686\tnote{**}} & 
    \new{0.567\tnote{**}} & 
    \new{0.602\tnote{**}} && 
    \new{0.653\tnote{**}} & 
    \new{0.527\tnote{**}} & 
    \new{0.561\tnote{**}} &&
    \new{0.697\tnote{*}} & 
    \new{0.565\tnote{*}} & 
    \new{0.613\tnote{*}} \\
    ROI 3& 
    \new{0.544\tnote{*}} & 
    \new{0.375\tnote{*}} & 
    \new{0.452\tnote{*}} && 
    \new{0.542\tnote{*}} & 
    \new{\textbf{0.388}\tnote{+}} & 
    \new{\textbf{0.458}\tnote{+}} &&
    \new{0.546\tnote{*}} & 
    \new{0.369\tnote{*}} & 
    \new{0.454\tnote{*}}\\
    ALL& 
    \new{0.669\tnote{*}} & 
    \new{0.518\tnote{**}} & 
    \new{0.618\tnote{**}} && 
    \new{0.651\tnote{**}} & 
    \new{0.495\tnote{**}} & 
    \new{0.599\tnote{**}} &&
    \new{0.665\tnote{**}}& 
    \new{0.509\tnote{**}}& 
    \new{0.615\tnote{**}}\\
    \midrule
    & \multicolumn{3}{l}{\tiny $s_w=0.25\,, s_h = 0.15$} & & \multicolumn{3}{l}{\tiny $s_w=0.25\,, s_h = 0.2$} & & \multicolumn{3}{l}{\tiny $s_w=0.25\,, s_h = 0.25$}\\
    \cmidrule{2-4}
    \cmidrule{6-8}
    \cmidrule{10-12}
    ROI 1& 
    \new{\textbf{0.782}\tnote{+}} & 
    \new{\textbf{0.668}\tnote{+}} & 
    \new{\textbf{0.678}\tnote{+}} && 
    \new{0.764\tnote{*}} & 
    \new{0.642\tnote{**}} & 
    \new{0.654\tnote{*}} && 
    \new{0.756\tnote{*}} & 
    \new{0.633\tnote{**}} & 
    \new{0.642\tnote{**}}\\
    ROI 2& 
    \new{\textbf{0.711}\tnote{+}} & 
    \new{\textbf{0.593}} & 
    \new{\textbf{0.630}\tnote{+}} &&
    \new{\textbf{0.709}\tnote{+}} & 
    \new{0.581\tnote{*}} & 
    \new{\textbf{0.626}\tnote{+}} && 
    \new{0.703\tnote{*}} & 
    \new{0.573\tnote{*}} &
    \new{\textbf{0.620}\tnote{+}} \\
    ROI 3& 
    \new{\textbf{0.558}} & 
    \new{\textbf{0.388}} & 
    \new{\textbf{0.466}} && 
    \new{\textbf{0.552}\tnote{+}} & 
    \new{\textbf{0.379}\tnote{+}} & 
    \new{\textbf{0.464}\tnote{+}} && 
    \new{0.542\tnote{*}} & 
    \new{\textbf{0.384}\tnote{+}} &
    \new{\textbf{0.459}\tnote{+}} \\
    ALL& 
    \new{\textbf{0.684}} & 
    \new{\textbf{0.535}\tnote{+}} & 
    \new{\textbf{0.635}} && 
    \new{0.675\tnote{*}} & 
    \new{0.521\tnote{**}} & 
    \new{0.626\tnote{**}} &&
    \new{0.667\tnote{**}}& 
    \new{0.511\tnote{**}}& 
    \new{0.617\tnote{**}}\\
    \midrule
    & \multicolumn{3}{l}{\tiny $s_w=0.5\,, s_h = 0.45$} & & \multicolumn{3}{l}{\tiny $s_w=0.5\,, s_h = 0.6$} & & \multicolumn{3}{l}{\tiny $s_w=0.5\,, s_h = 0.75$}\\
    \cmidrule{2-4}
    \cmidrule{6-8}
    \cmidrule{10-12}
    ROI 1& 
    \new{\textbf{0.786}} & 
    \new{\textbf{0.680}} & 
    \new{\textbf{0.684}} && 
    \new{0.758\tnote{**}} & 
    \new{0.641\tnote{**}} & 
    \new{0.646\tnote{**}} && 
    \new{0.745\tnote{**}} & 
    \new{0.626\tnote{**}} & 
    \new{0.629\tnote{**}} \\
    ROI 2 &     
    \new{\textbf{0.714}} & 
    \new{\textbf{0.589}\tnote{+}} & 
    \new{\textbf{0.631}} &&
    \new{0.697\tnote{*}} & 
    \new{0.563\tnote{**}} & 
    \new{0.610\tnote{*}} &&
    \new{0.697\tnote{*}} & 
    \new{0.571\tnote{*}} & 
    \new{0.614\tnote{*}} \\
    ROI 3 & 
    \new{0.534\tnote{**}} &
    \new{\textbf{0.380}\tnote{+}} & 
    \new{0.447\tnote{*}} && 
    \new{0.524\tnote{**}} & 
    \new{0.370\tnote{*}} & 
    \new{0.439\tnote{**}} && 
    \new{0.520\tnote{**}} & 
    \new{0.364\tnote{*}} & 
    \new{0.438\tnote{**}} \\
    ALL& 
    \new{\textbf{0.678}\tnote{+}} & 
    \new{\textbf{0.539}} & 
    \new{\textbf{0.629}\tnote{+}} && 
    \new{0.659\tnote{**}}& 
    \new{0.510\tnote{**}}& 
    \new{0.609\tnote{**}}&&
    \new{0.654\tnote{**}}& 
    \new{0.496\tnote{**}}& 
    \new{0.603\tnote{**}}\\
    \bottomrule
    \bottomrule
    \end{tabular}
    \begin{tablenotes}
         \item[+] $\ p\geq0.05$; $^{*}\ p<0.05$; $^{**}\ p<0.001$; unpaired t-test with respect to top result.
    \end{tablenotes}
    \end{threeparttable}
      \caption*{Classification performance for different $s_w$, $s_h$ values on the \ac{ROIs}.}
    \end{minipage}
  \begin{minipage}[b]{0.29\textwidth}
    \centering
    \includegraphics[width=\linewidth]{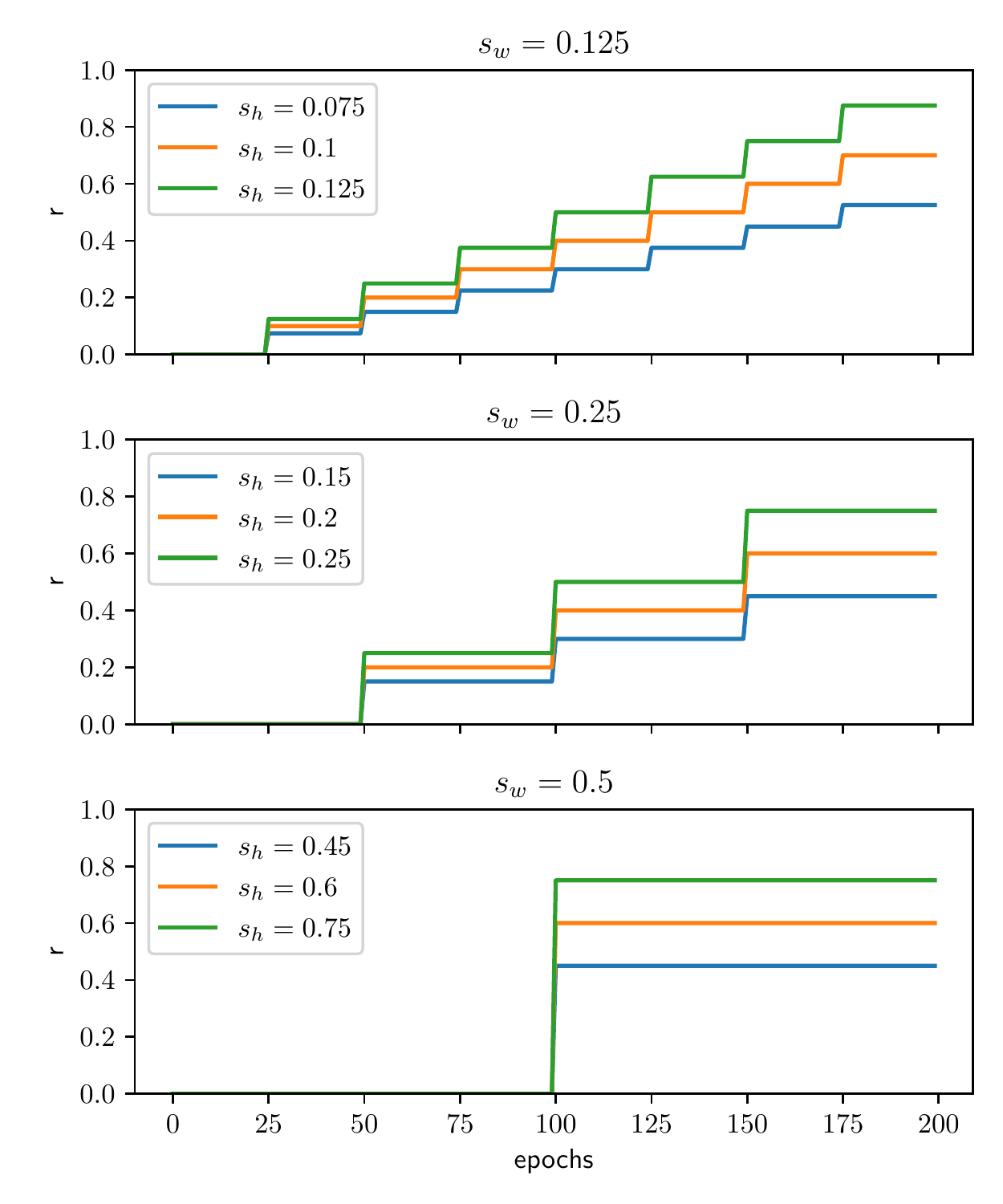} 
    \caption*{Profile of the \acl{E2H} ratio $r$.}
  \end{minipage}
 \caption{Importance of $s_w$ and $s_h$ parameter tuning for the \acl{E2H} learning scheme. (left) Performance of the model on the three \acf{ROIs} for each parameter pair. (right) Corresponding profiles of the step function $r$ (Equation~\ref{eq:r}) as a function of the current epoch. The variable $r$ represents the fraction of the "trusted" samples included for cross-domain matching, based the cross-entropy.}
   \label{fig:e2h_sweep}
  \end{figure}

Training with using only $\mathcal{L}_{\mathrm{IND}}$ achieves relatively good performances but fails to generalize knowledge to classes where textures and staining strongly vary.
\new{In the patch classification task for example, this is apparent for the background and tumor class.
For the second evaluation task, we can observe the same trend in the ROI $3$ where tumor and normal stroma are mixed.}

Using only $\mathcal{L}_{\mathrm{CRD}}$ does not help and creates an unstable model. As we do not impose domain representation, the model converges toward incorrect solutions where random sets of samples are matched between the source and target datasets. Moreover, it can create degenerated solutions where examples from the source and target domain are perfectly matched even though they do not present any visual similarity. \new{The reformulation of the entropy, however, slightly improves the cross domain matching.}

Even the combination of the in-domain and cross-domain loss is not sufficient to improve the capability of the model. When performing a class-wise analysis, we observe that the performance on tumor and debris detection drastically dropped without the entropy reformulation. Both classes are forced to match samples from other classes, thus worsening the representation of the embedding.

The introduction of the \ac{E2H} procedure improves the \new{overall classification as well as most of the per-class performance for the first task. In the second task, it improves the performance across all metrics in all three \ac{ROIs}.} 
The importance of the \ac{E2H} learning is evaluated and discussed in more detail in the next section.


\new{Overall, the updated definition of the entropy improves the model's performance for both the cross-domain patch classification and \ac{WSI}s segmentation task.
It helps to ensure that both model branches output a similar distribution, thus providing better cross-domain candidates. The improvement is most visible for the tumor and stroma predictions.}

\begin{figure*}[!htb]
\centering
  \includegraphics[width=0.99\textwidth]{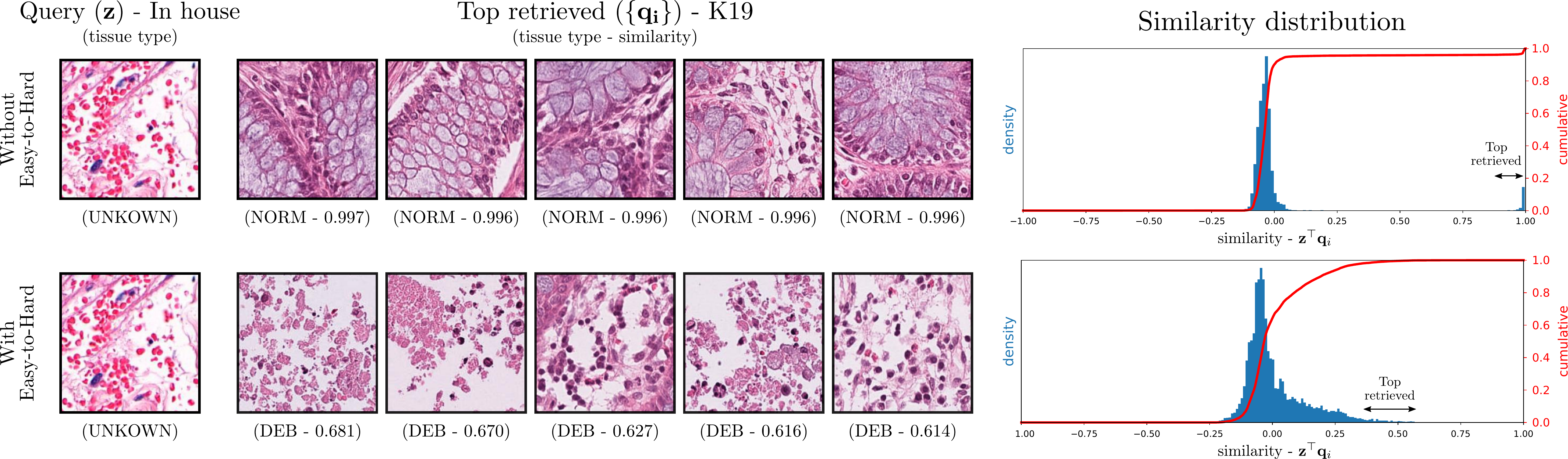}
  \caption{\new{Importance of the \acf{E2H} learning scheme for the cross-domain image retrieval. The first column shows the input query image $\mathbf{z}$ from our in-house cohort (target domain)}, the second column presents the retrieved samples from \ac{K19} that have the highest similarity in the source queue $\{\mathbf{q}_i\}$, and the third column shows the density distribution (blue) of similarities across the source queue as well as its cumulative profile (red). We list the retrieved examples with their assigned classes. The query class is unknown. 
  The top and bottom rows highlight the result of training without and with \ac{E2H} learning, respectively.
  Without \ac{E2H}, the model tries to optimize $\mathcal{L}_{\mathrm{CRD}}$ at any cost, which creates out-of-distribution samples (seen at the very right).
  With \ac{E2H} the model predicts samples with lower confidence, but that are still visually similar.
  }
  \label{app:query}
\end{figure*}

\subsection{Evaluation of the \ac{E2H} Learning Scheme}
\label{subsec:e2h}

In this section, we discuss the usefulness and robustness of the \ac{E2H} learning. The learning procedure is based on $r$, and the two contributing variables $s_w$ and $s_h$: \new{

\begin{equation}
    r = \Big\lfloor \frac{e}{N_\mathrm{epochs} \cdot s_{w}} \Big\rfloor \cdot s_{h}, \tag{\ref{eq:r} revisited}
\end{equation}
}

In Figure~\ref{fig:e2h_sweep}, we show the impact of different combinations of these parameters on the single cross-domain segmentation task (see Section~ \ref{subsec:crossdomain_seg}).
We report the pixel-wise accuracy, the weighted \acl{IoU}, and the pixel-wise Cohen's kappa ($\kappa$) score for the presented \ac{ROIs}. For each parameter pair, we also display the profile of the variable $r$ as a function of the of the epoch $e$.

Firstly, we observe that the model is more robust when $s_h$ is low. The variable is an indicator of the ratio of samples used for cross-domain matching. In other words, the architecture benefits from a small $s_h$ that allows it to focus on examples with high similarity/confidence while avoiding complex samples without properly matching candidates.
Secondly, the selection of $s_w$ is also crucial to the stability of the prediction. 
This quantity measures the number of epochs to wait before considering more complex examples in the cross-domain matching optimization.
For small $s_w$ values, the model has no time to learn the feature representation properly before encountering more difficult samples. This is especially true for the first few epochs after initialization, where the architecture is not yet able to optimally embed features. Furthermore, using large $s_w$ weakens the model capability to progressively learn from more complex samples.

Figure~\ref{app:query} shows an example patch from the training phase and highlights the usefulness of the \ac{E2H} scheme. When dealing with a heterogeneous target data cohort, some tissue types might not have relevant candidates in the other set (open-set scenario). The presented example shows an example composed of a vein and blood cells. 
Such a tissue structure is absent from the source cohort, and thus does not have matching sample in the target domain.

Without the \ac{E2H} learning, the model is forced to find matching candidates for the query $\mathbf{z}$, here normal mucosa (NORM), to minimize the cross-entropy term $\bar{H}$. 
When plotting the similarity distribution, the matched samples form an out-of-distribution cluster with a high similarity to the query ($\mathbf{z}^{\top} \mathbf{q}_i \simeq 1)$.  
This phenomenon is even more visible with the cumulative function (red) that tends to the step function. 

When training with the \ac{E2H} scheme, we observe a continuous transition in the distribution of samples similarities. Here, the top retrieved samples share the same granular structure as the query. Still, we have to be careful as they do not represent the same type of tissue. The retrieved samples are examples of necrosis, whereas the query shows red blood cells. 
The fact that the architecture is less confident (i.e., the similarity is lower for the top retrieved samples) is a good indicator of its robustness and ability to process complex queries.

As a result, the introduction of the \ac{E2H} process prevents the model from learning degenerated solutions. 
We also observe this with other open-set tissue classes such as complex stroma and loose connective tissue, which are absent in the source domain.

\begin{table*}[ht]
    \caption{\new{Performance of the \acf{SRMA} framework on the \ac{CRCTP} dataset in a multi-source domain setting. We show the results for different combinations of \ac{K16} and \ac{K19} used for the self-supervised pre-training as well as training the classification header. For the source domains \ac{K19} and \ac{K16}, only $1\%$ and $10\%$ of the labeled data are used, respectively. 
    We  also compare the performance of the $1:1$ with the $K:1$ setting for the loss definitions (see Equations~\ref{eq:L_ind_11}-\ref{eq:L_crd_K1}). 
    We report the F1 score for the individual classes and weighted F1 score for the overall mean performance (all) (averaged over 10 runs). Some classes have been merged due to overlapping definitions.
    }}
    \label{tbl:k16-k19-crctp-classification}
    \centering
    \scriptsize
    \begin{threeparttable}
    \begin{tabular}{p{0.2cm}lcccccccccrrrrrrr}
    \toprule
    \toprule
    &&\multicolumn{2}{c}{Pretraining} && \multicolumn{2}{c}{Classification} && \multicolumn{2}{c}{Multi-source} \\
    \cmidrule{3-4}
    \cmidrule{6-7}
    \cmidrule{9-10}
    \multicolumn{2}{l}{Methods} & K19 & K16 && K19 & K16 &&  $\mathcal{L}_{\textrm{IND}}$ & $\mathcal{L}_{\textrm{CRD}}$ && TUM & STR\tnote{$\dagger$} & LYM & NORM & DEB\tnote{$\dagger$} & ALL\\
    \midrule
    \multicolumn{2}{l}{\new{\textit{Single source:}}} \\
    \cmidrule{2-17}
    &\new{SRA \citep{abbet2021selfrule}} & \new{-} & \new{$\checkmark$} && \new{-} & \new{$\checkmark$} && - & - &&
    \new{\textbf{82.2}} & \new{\textbf{69.3}} & \new{62.5\tnote{**}} & \new{\textbf{69.8}} & \new{47.4\tnote{*}} & \new{\textbf{69.4}}
    \\
    &\new{SRMA} & \new{-} & \new{$\checkmark$} && \new{-} & \new{$\checkmark$} && - & - &&
    \new{\textbf{82.0}\tnote{+}} & \new{63.5\tnote{**}} & \new{\textbf{66.3}} & \new{51.9\tnote{**}} & \new{\textbf{50.3}} & \new{65.2\tnote{**}}
    \\
    \cmidrule{2-17}
    &\new{SRA \citep{abbet2021selfrule}} & \new{$\checkmark$} & \new{-} && \new{$\checkmark$} & \new{-} && - & - && 
    \new{91.0\tnote{*}} & \new{84.9\tnote{**}} & \new{62.0\tnote{**}} & \new{\textbf{71.7}} & \new{\textbf{58.5}\tnote{+}} & \new{79.2\tnote{**}}
    \\
    &\new{SRMA} & \new{$\checkmark$} & \new{-} && \new{$\checkmark$} & \new{-} && - & - &&
    \new{\textbf{91.7}} & \new{\textbf{86.7}} & \new{\textbf{65.4}} & \new{68.6\tnote{**}} & \new{\textbf{58.9}} & \new{\textbf{80.2}}
    \\
    \cmidrule{2-17}
    \\
    \multicolumn{2}{l}{\new{\textit{Multi source:}}} \\
    \cmidrule{2-17}
    &\new{DeepAll \citep{dou2019domain}} & \new{$\checkmark$} & \new{$\checkmark$} && \new{-} & \new{$\checkmark$} &&  - & - &&
    \new{52.4\tnote{**}} & \new{64.1\tnote{**}} & \new{36.5\tnote{**}} & \new{14.2\tnote{**}} & \new{13.8\tnote{**}} & \new{47.1\tnote{**}}
    \\
    &\new{SRA \citep{abbet2021selfrule}} & \new{$\checkmark$} & \new{$\checkmark$} && \new{-} & \new{$\checkmark$} && \new{$1:1$} & \new{$1:1$} && \new{70.9\tnote{**}} & \new{68.5\tnote{**}} & \new{45.6\tnote{**}} & \new{72.2\tnote{**}} & \new{19.1\tnote{**}}  & \new{62.2\tnote{**}}
    \\
    &\new{SRMA} & \new{$\checkmark$} & \new{$\checkmark$} && \new{-} & \new{$\checkmark$} && \new{$1:1$} & \new{$1:1$} && \new{76.6\tnote{**}} & \new{69.3\tnote{**}} & \new{48.7\tnote{**}} & \new{74.5\tnote{**}} & \new{18.2\tnote{**}}  & \new{64.4\tnote{**}}
    \\
    &\new{SRMA} & \new{$\checkmark$} & \new{$\checkmark$} && \new{-} & \new{$\checkmark$} && \new{$K:1$} & \new{$1:1$} &&
    \new{\textbf{89.4}\tnote{+}} & \new{\textbf{74.9}\tnote{+}} & \new{\textbf{66.8}} & \new{\textbf{75.6}} & \new{\textbf{43.7}} & \new{\textbf{74.4}}
    \\
    &\new{SRMA} & \new{$\checkmark$} & \new{$\checkmark$} && \new{-} & \new{$\checkmark$} && \new{$1:1$} & \new{$K:1$} && \new{75.9\tnote{**}} & \new{73.3\tnote{*}} & \new{45.9\tnote{**}} & \new{73.0\tnote{**}} & \new{22.6\tnote{**}}  & \new{65.8\tnote{**}}
    \\
    &\new{SRMA} & \new{$\checkmark$} & \new{$\checkmark$} && \new{-} & \new{$\checkmark$} && \new{$K:1$} & \new{$K:1$} && 
    \new{\textbf{89.8}} & \new{\textbf{75.2}} & \new{64.5\tnote{**}} & \new{74.1\tnote{**}} & \new{25.7\tnote{**}} & \new{72.5\tnote{**}}
    \\
    \cmidrule{2-17}
    &\new{DeepAll \citep{dou2019domain}} & \new{$\checkmark$} & \new{$\checkmark$} && \new{$\checkmark$} & \new{-} && - & - && 
    \new{72.4\tnote{**}} & \new{88.6\tnote{**}} & \new{43.6\tnote{**}} & \new{53.2\tnote{**}} & \new{71.8\tnote{**}} & \new{73.2\tnote{**}}
    \\
    &\new{SRA \citep{abbet2021selfrule}} & \new{$\checkmark$} & \new{$\checkmark$} && \new{$\checkmark$} & \new{-} &&\new{$1:1$} & \new{$1:1$} && \new{86.2\tnote{**}} & \new{87.6\tnote{**}} & \new{66.7\tnote{**}} & \new{71.0\tnote{**}} & \new{\textbf{80.5}} & \new{81.8\tnote{**}}
    \\
    &\new{SRMA} & \new{$\checkmark$} & \new{$\checkmark$} && \new{$\checkmark$} & \new{-} && \new{$1:1$} & \new{$1:1$} && \new{\textbf{92.5}} & \new{88.4\tnote{**}} & \new{68.7\tnote{**}} & \new{68.3\tnote{**}} & \new{74.2\tnote{*}} & \new{82.9\tnote{*}}
    \\
    &\new{SRMA} & \new{$\checkmark$} & \new{$\checkmark$} && \new{$\checkmark$} & \new{-} && \new{$K:1$} & \new{$1:1$} &&
    \new{91.5\tnote{*}} & \new{87.6\tnote{**}} & \new{\textbf{70.7}} & \new{\textbf{75.0}} & \new{65.7\tnote{**}} & \new{82.7\tnote{*}}
    \\
    &\new{SRMA} & \new{$\checkmark$} & \new{$\checkmark$} && \new{$\checkmark$} & \new{-} && \new{$1:1$} & \new{$K:1$} && \new{90.1\tnote{**}} & \new{\textbf{90.1}} & \new{\textbf{69.6}\tnote{+}} & \new{72.9\tnote{**}} & \new{71.6\tnote{**}} & \new{\textbf{83.6}}
    \\
    &\new{SRMA} & \new{$\checkmark$} & \new{$\checkmark$} && \new{$\checkmark$} & \new{-} && \new{$K:1$} & \new{$K:1$} &&
    \new{\textbf{91.6}\tnote{+}} & \new{87.4\tnote{**}} & \new{68.7\tnote{**}} & \new{73.9\tnote{**}} & \new{53.3\tnote{**}} & \new{81.2\tnote{**}}
    \\
    \cmidrule{2-17}
    &\new{DeepAll \citep{dou2019domain}} & \new{$\checkmark$} & \new{$\checkmark$} && \new{$\checkmark$} & \new{$\checkmark$} && - & - &&
    \new{81.4\tnote{**}} & \new{\textbf{85.7}\tnote{+}} & \new{50.9\tnote{**}} & \new{50.1\tnote{**}} & \new{51.5\tnote{**}} & \new{72.6\tnote{**}}
    \\
    &\new{SRA \citep{abbet2021selfrule}} & \new{$\checkmark$} & \new{$\checkmark$} && \new{$\checkmark$} & \new{$\checkmark$}&& \new{$1:1$} & \new{$1:1$} && \new{85.8\tnote{**}} & \new{\textbf{85.9}} & \new{72.9\tnote{*}} & \new{72.1\tnote{**}} & \new{\textbf{59.2}} & \new{\textbf{80.1}}
    \\
    &\new{SRMA} & \new{$\checkmark$} & \new{$\checkmark$} && \new{$\checkmark$} & \new{$\checkmark$} && \new{$1:1$} & \new{$1:1$} && \new{\textbf{92.9}} & \new{82.4\tnote{**}} & \new{72.1\tnote{*}} & \new{70.8\tnote{**}} & \new{53.7\tnote{**}} & \new{79.3\tnote{*}}
    \\
    &\new{SRMA} & \new{$\checkmark$} & \new{$\checkmark$} && \new{$\checkmark$} & \new{$\checkmark$} && \new{$K:1$} & \new{$1:1$} &&
    \new{\textbf{92.8}\tnote{+}} & \new{81.7\tnote{**}} & \new{\textbf{73.5}} & \new{\textbf{74.6}} & \new{49.8\tnote{**}} & \new{79.3\tnote{*}}
    \\
    &\new{SRMA} & \new{$\checkmark$} & \new{$\checkmark$} && \new{$\checkmark$} & \new{$\checkmark$} && \new{$1:1$} & \new{$K:1$} && \new{89.6\tnote{**}} & \new{84.7\tnote{*}} & \new{72.5\tnote{*}} & \new{\textbf{74.4}\tnote{+}} & \new{52.1\tnote{**}} & \new{\textbf{80.0}\tnote{+}}
    \\
    &\new{SRMA} & \new{$\checkmark$} & \new{$\checkmark$} && \new{$\checkmark$} & \new{$\checkmark$} && \new{$K:1$} & \new{$K:1$} &&
    \new{92.5\tnote{*}} & \new{80.6\tnote{**}} & \new{70.5\tnote{**}} & \new{73.9\tnote{**}} & \new{39.4\tnote{**}} & \new{77.4\tnote{**}}
    \\
    \bottomrule
    \bottomrule
    \end{tabular}
    \begin{tablenotes}
        \item[$\dagger$] \new{The STR and MUS classes are merged as STR class; DEB and MUC classes as DEB.}
        \item[+] $\ p\geq0.05$; $^{*}\ p<0.05$; $^{**}\ p<0.001$; unpaired t-test with respect to top result.
    \end{tablenotes}
    \end{threeparttable}
\end{table*}

\subsection{\new{Multi-Source Patch Classification}}
\label{subsec:multi_source_patch}

\new{
We explore the benefit of using multiple source domains with different distributions to perform domain adaptation for the patch classification task. 
To do so, we select \ac{K19} and \ac{K16} as the source sets and \ac{CRCTP} as the target set. 
To learn the feature representations, the model is trained in an unsupervised fashion using both source domains as well as the unlabeled target domain. 
For the evaluation, we train a linear classifier on top of the frozen features with few randomly selected labeled samples from the source domains (1000 samples from \ac{K19} (1\%), and 500 samples from \ac{K16} (10\%)). 
By using only little labeled data, we aim to reduce the annotation workload for pathologists while still achieving good classifications performances. The set of labeled data differs between each run, as they are randomly sampled for each individual run.

The three datasets \ac{K19}, \ac{K16}, and \ac{CRCTP} do not have one-to-one classes correspondence. Thus, for the evaluation of the target set, we only consider the classes present in all datasets, namely, tumor (TUM), stroma (STR), lymphocytes (LYM), normal mucosa (NORM), and debris / necrotic tissue (DEB).
Still, during the unsupervised pre-training we consider all classes, including those who do not have matching candidates across the sets, such as background (BACK) and adipose (ADI).
This setup creates an open-set scenario for the cross-domain matching and allows the model to learn more robust features representations. 

For comparison purposes, we use the same hyper-parameters as in the single source domain patch classification setting with $s_w=0.25$, $s_h=0.15$. 
The probability of drawing a sample $\textbf{x}$ from the source or the target domain is the same and is given by $p(\textbf{x} \in \mathcal{D}_s) = K p(\textbf{x} \in \mathcal{D}_s^k) = p(\textbf{x} \in \mathcal{D}_t)$, where K is the number of source domains.
}

\new{
The results are presented in Table~\ref{tbl:k16-k19-crctp-classification}. We compare the performance of different experimental setups in regards to the used datasets and multi-source scenario for our \ac{SRMA}.
We show three scenarios where we use either \ac{K16}, \ac{K19}, or the combination of the two (\ac{K16} and \ac{K19}) to train the classification layer.
To evaluate the impact of the multi-source scenario, where we investigate all possibilities for the in-domain ($\mathcal{L}_{\mathrm{IND}}^{1:1}, \mathcal{L}_{\mathrm{IND}}^{K:1}$) and cross-domain ($\mathcal{L}_{\mathrm{CRD}}^{1:1}, \mathcal{L}_{\mathrm{CRD}}^{K:1}$) loss definitions, as introduced in Equations~\ref{eq:L_ind_11}-\ref{eq:L_crd_K1}).
As baselines, we consider the single source setting of the presented \ac{SRMA} model, our previous \ac{SRA} work, as well as the DeepAll approach that uses aggregation of all the source tissue data into a single training set \citep{dou2019domain}.
}

\new{
The \ac{SRMA} and \ac{SRA} single source baselines both show a better performance for \ac{K19} compared to \ac{K16}.
This is most likely due to the fact that the \ac{K16} subset for training the classification header is relatively small, with only $5,000$ different examples.
Also, \ac{SRMA} outperforms our previous \ac{SRA} work for all classes except one, which is an indicator of the robustness of the entropy reformulation.
}

\new{
For the multi-source adaptation, we show three scenarios where we use either \ac{K16}, \ac{K19}, or the combination of the two (\ac{K16} and \ac{K19}) to train the classification layer.
When using solely \ac{K16}, we can observe that the debris classification tends to have lower performances across all models. Debris examples in \ac{K16} appear highly saturated, which makes the generalization of the class a challenging task.
Only the proposed \ac{SRMA} approach is able to achieve better performances compared to the single source baselines. 
Using \ac{K19} for the classification of target patches gives overall the best performance. 
Interestingly, using both \ac{K19} and \ac{K16} leads to a drop in performance. This is most likely due to potential discrepancies between the class definitions, which makes it more difficult for the model to generalize the class representations across the different modalities.
}

\new{
When comparing the in-domain and cross-domain multi-source scenarios, we find that using $\mathcal{L}_{\textrm{IND}}^{1:1}$ and $\mathcal{L}_{\textrm{CRD}}^{K:1}$ achieves the best results across the various settings.
This suggests that it is better to optimize the source domain as a single set for the in-domain representation. 
However, when performing cross-domain matching, considering domain to domain correspondence between each source set and the target domain yields better performances.
It ensures that the model looks for relevant candidates in all individual source sets as tissue samples might have a distinct appearance in different source domains.

We also note that $\mathcal{L}_{\textrm{IND}}^{K:1}$ is only relevant when only using \ac{K16} to train the classification header.
This is due to the fact that the cross-domain matching fails to retrieve debris samples correctly from the \ac{K16} domain, which tend to be misclassified as lymphocytes because of their similar granular appearance and as well as their hematoxylin-positive aspect.
Overall the combination of both $\mathcal{L}_{\textrm{IND}}^{K:1}$ and $\mathcal{L}_{\textrm{CRD}}^{K:1}$ degrades the performance slightly. 

Complementary results on the importance of the dataset ratios when sampling data for the unsupervised pre-training phase are available in \ref{app:bs_sampling}.
}

\subsection{Use Case: Multi-source Segmentation of \ac{WSI}}
\label{subsec:multi_source_wsi}

In this section, we present the results for the multi-source domain adaptation for patch-based segmentation of \ac{WSI} \ac{ROIs}. 
More specifically, we are interested in the detection of desmoplastic reactions (complex stroma), which is a prognostic factor in \ac{CRC} \citep{ueno2021histopathological}.
We use both \ac{K19} and \ac{CRCTP} as the source datasets to add complex stroma examples to the source domain. Our in-house dataset is used as the target domain. 

To assess the quality of the prediction, we evaluate the models on the same \ac{ROIs} as in the single-source setting. 
However, the previously provided annotations do not include complex stroma. 
We overcome this by defining a margin around the tumor tissue in the existing annotations, which is considered as the interaction area. Stroma in this region is therefore re-annotated as complex stroma. 
The margin is fixed to $500\mu m$ such that it includes the close tumor neighborhood and matches the definition of complex stroma in the literature \citep{berben2020computerised, nearchou2021automated}.

\begin{table*}[ht]
    \caption{
    \new{
    Analysis of the performance of the \acf{SRMA} approach in regards to complex stroma detection.
    Multiple possible scenarios are evaluated in regard to the data included for pre-training, as well as the multi-source setting ($1:1$ versus $K:1$, see  Equations~\ref{eq:L_ind_11}-\ref{eq:L_crd_K1}), as indicated in the table. Only $1\%$ of the labels are used for the classification stage.
    We report the F1-score for complex stroma, the overall weighed F1-score, the pixel-wise accuracy, the dice score, the weighted intersection over union (IoU), and the pixel-wise Cohen’s kappa ($\kappa$) (averaged over 10 runs).
    }
    }
    \label{tbl:k19-crctp-wsi-classification}
    \centering
    \scriptsize
    \begin{threeparttable}
    \begin{tabular}{llcccccrrrrrrr}
    \toprule
    \toprule
    && \multicolumn{2}{c}{Pretraining} && \multicolumn{2}{c}{Multi-source} && 
    \\
    \cmidrule{3-4}
    \cmidrule{6-7}
    \multicolumn{2}{l}{Model} & K19 & CRCTP && $\mathcal{L}_{\textrm{IND}}$ & $\mathcal{L}_{\textrm{CRD}}$ && F1-CSTR\tnote{$\dagger$} & F1-ALL & Acc. & Dice & IoU & $\kappa$ \\
    \midrule
    \multicolumn{2}{l}{\new{\textit{ROI 1-3 (w/o CSTR)}}} \\
    \cmidrule{2-14}
    & \new{DeepAll \citep{dou2019domain}} & \new{$\checkmark$} & \new{$\checkmark$} && \new{-} &  \new{-} && \new{-} & \new{0.622\tnote{**}}  & \new{0.615\tnote{**}} & \new{0.583\tnote{**}} & \new{0.483\tnote{**}} & \new{0.552\tnote{**}}
    \\ 
    &\new{SRA \citep{abbet2021selfrule}} & \new{$\checkmark$} & \new{-} && \new{-} & \new{-} && \new{-} & \new{0.648\tnote{**}} & \new{0.661\tnote{**}} & \new{0.632\tnote{**}} & \new{0.521\tnote{**}} & \new{0.611\tnote{**}}
    \\
    &\new{SRMA} & \new{$\checkmark$} & \new{-} && \new{-} & \new{-} && \new{-} & \new{\textbf{0.667}\tnote{+}} & \new{\textbf{0.684}\tnote{+}} & \new{0.647\tnote{**}} & \new{\textbf{0.536}\tnote{+}} & \new{\textbf{0.636}\tnote{+}}
    \\
    &\new{SRMA} & \new{$\checkmark$} & \new{$\checkmark$} &&\new{$1:1$} & \new{$1:1$} && \new{-} & \new{\textbf{0.673}} & \new{\textbf{0.685}} & \new{\textbf{0.669}} & \new{\textbf{0.541}} & \new{\textbf{0.636}}
    \\
    &\new{SRMA} & \new{$\checkmark$} & \new{$\checkmark$} &&\new{$K:1$} & \new{$1:1$} && \new{-} & \new{0.644\tnote{**}} & \new{0.665\tnote{**}} & \new{0.637\tnote{**}} & \new{0.516\tnote{**}} & \new{0.615\tnote{**}}
    \\
    &\new{SRMA} & \new{$\checkmark$} & \new{$\checkmark$} &&\new{$1:1$} & \new{$K:1$} && \new{-} & \new{\textbf{0.662}\tnote{+}} & \new{\textbf{0.678}\tnote{+}} & \new{0.652\tnote{*}} & \new{0.528\tnote{*}} & \new{\textbf{0.629}\tnote{+}}
    \\
    &\new{SRMA}& \new{$\checkmark$} & \new{$\checkmark$} &&\new{$K:1$} & \new{$K:1$} && \new{-} & \new{0.638\tnote{**}} & \new{0.660\tnote{**}} & \new{0.632\tnote{**}} & \new{0.509\tnote{**}} & \new{0.609\tnote{**}}
    \\
    \\
    \multicolumn{2}{l}{\new{\textit{ROI 1-3 (w/ CSTR)}}} \\
    \cmidrule{2-14}
    & \new{DeepAll \citep{dou2019domain}} & \new{$\checkmark$} & \new{$\checkmark$} && \new{-} & \new{-} && \new{0.001\tnote{**}} & \new{0.505\tnote{**}} & \new{0.539\tnote{**}} & \new{0.496\tnote{**}} & \new{0.399\tnote{**}} & \new{0.479\tnote{**}}
    \\
    &\new{SRA \citep{abbet2021selfrule}} & \new{$\checkmark$} & \new{-} && \new{-} & \new{-} && \new{0.214\tnote{**}} & \new{0.600\tnote{**}} & \new{0.624\tnote{**}} & \new{0.582\tnote{**}} & \new{0.490\tnote{**}} & \new{0.577\tnote{**}}
    \\
    &\new{SRMA} & \new{$\checkmark$} & \new{-} && \new{-} & \new{-} && \new{0.263\tnote{**}} & \new{0.614\tnote{**}} & \new{0.641\tnote{**}} & \new{0.595\tnote{**}} & \new{0.498\tnote{**}} & \new{0.594\tnote{**}}
    \\
    &\new{SRMA}& \new{$\checkmark$} & \new{$\checkmark$} && \new{$1:1$} & \new{$1:1$} && \new{\textbf{0.479}\tnote{+}} & \new{\textbf{0.647}\tnote{+}} & \new{0.659\tnote{*}} & \new{\textbf{0.631}} & \new{\textbf{0.524}\tnote{+}} & \new{0.613\tnote{**}}
    \\
    &\new{SRMA} & \new{$\checkmark$} & \new{$\checkmark$} && \new{$K:1$} & \new{$1:1$} && \new{\textbf{0.492}} & \new{\textbf{0.650}} & \new{\textbf{0.669}} & \new{0.618\tnote{**}} & \new{\textbf{0.524}} & \new{\textbf{0.624}}
    \\
    &\new{SRMA}& \new{$\checkmark$} & \new{$\checkmark$} && \new{$1:1$} & \new{$K:1$} && \new{\textbf{0.464}\tnote{+}} & \new{\textbf{0.640}\tnote{+}} & \new{0.651\tnote{**}} & \new{0.619\tnote{*}} & \new{0.513\tnote{*}} & \new{0.604\tnote{**}}
    \\
    &\new{SRMA}& \new{$\checkmark$} & \new{$\checkmark$} && \new{$K:1$} & \new{$K:1$} && \new{0.366\tnote{**}} & \new{0.623\tnote{**}} & \new{0.646\tnote{**}} & \new{0.597\tnote{**}} & \new{0.500\tnote{**}} & \new{0.599\tnote{**}}
    \\
    \bottomrule
    \bottomrule
    \end{tabular}
    \begin{tablenotes}
        \item[$\dagger$] \new{Performances are only available with extended annotations (w/CSTR).}
        \item[+] $\ p\geq0.05$; $^{*}\ p<0.05$; $^{**}\ p<0.001$; unpaired t-test with respect to top
    \end{tablenotes}
    \end{threeparttable}
\end{table*}

\begin{figure*}[t]
\centering
  \includegraphics[width=1.0\textwidth]{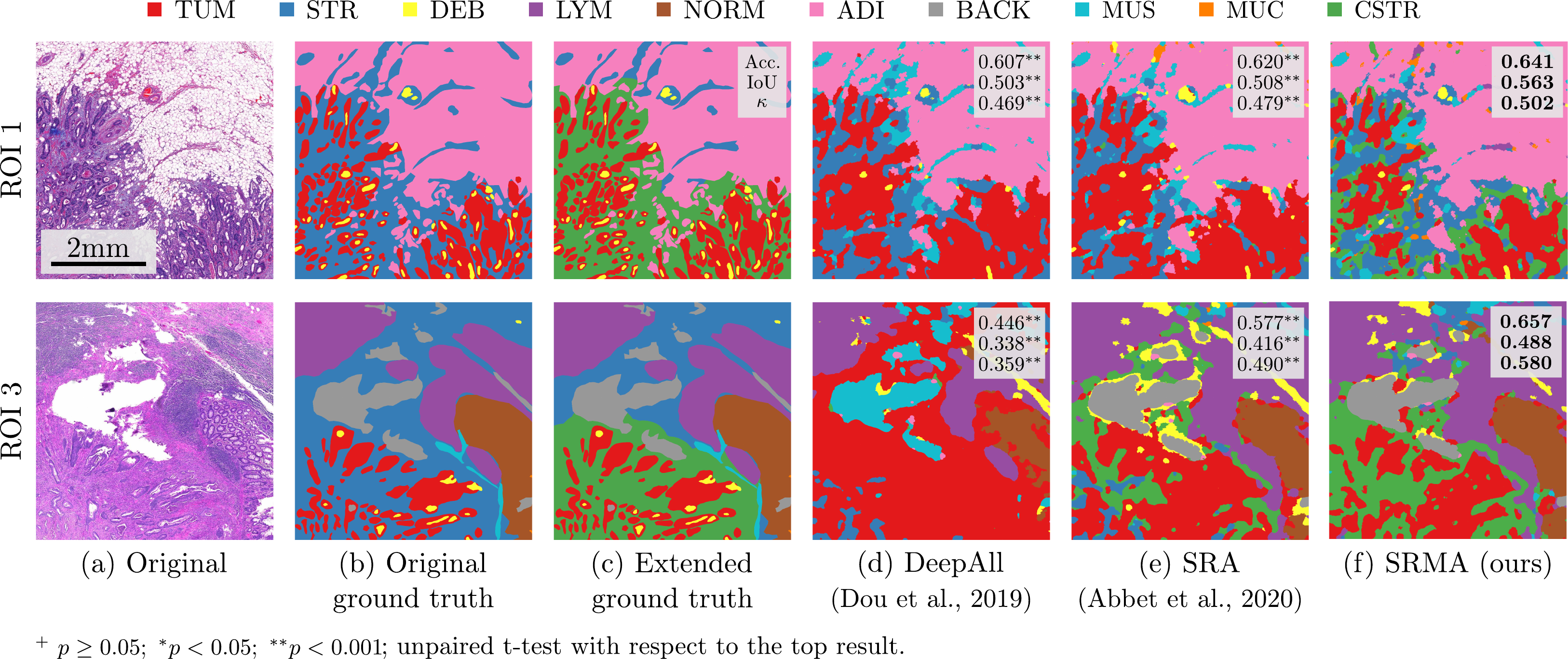}
  \caption{Results of the multi-source domain adaptation from \ac{K19} and \ac{CRCTP} to our in-house dataset. (a-c) show the original \acf{ROIs} from the \acp{WSI}, their original ground truth (without CSTR), and the extended ground truth (with CSTR), respectively. We compare the performance of our \ac{SRMA} \new{framework (f) to our previous work \ac{SRA} (e) and to the DeepAll baseline (d). For the multi-source optimization, we use the $1:1$ and $K:1$ approach for the in-domain and cross-domain, respectively}. We report the pixel-wise accuracy, the weighted \acl{IoU}, and the pixel-wise Cohen's kappa ($\kappa$) score averaged over 10 runs.}
  \label{fig:multisrc_cstr}
\end{figure*}

As a baseline, we use DeepAll, which aggregates all the source tissue data into a single training set \citep{dou2019domain}. The model is trained in an unsupervised fashion using a standard contrastive loss to optimize the data representation of the features \citep{chen2020improved}. 
In this case, no domain adaption is performed across the sets.

\new{
The results are presented in Table~\ref{tbl:k19-crctp-wsi-classification} and Figure~\ref{fig:multisrc_cstr}.
In Table~\ref{tbl:k19-crctp-wsi-classification}, we compare the performance of the models with and without complex stroma detection across all three \ac{ROIs}. 
We compare the single as well as the multi-source \ac{SRMA} approaches to the baselines, DeepAll and our previously published \ac{SRA} method. 
We report the F1-score for complex stroma, the overall weighted F1-score, the pixel-wise accuracy, the Dice score, the weighted intersection over union (IoU), and pixel-wise Cohen’s kappa ($\kappa$).

Without considering the complex stroma class, the numerical results show that all the multi-source settings achieve similar performances. 
Including an additional dataset, namely \ac{CRCTP}, does not improve nor seriously deteriorate the classification performances on the \ac{ROIs}.
Furthermore, merging the source domains for in-domain optimization ($\mathcal{L}_{\textrm{IND}}^{1:1}$) seems to be the best setup. 
For the cross-domain matching, both $\mathcal{L}_{\textrm{CRD}}^{1:1}$ and $\mathcal{L}_{\textrm{CRD}}^{K:1}$ achieve similar scores.

However, the benefit of using the multi-source approach can be observed when including complex stroma detection.
Here, the models which use \ac{CRCTP} as source set achieve better results. 
The detection of complex stroma improves by up to $20-25\%$. 
By contrast, the cross-domain matching on each subsets $\mathcal{L}_{\textrm{CRD}}^{K:1}$ penalizes the complex stroma detection. 
This can be explained by the fact that only \ac{CRCTP} contains examples of complex stroma. Therefore, imposing complex stroma retrieval in \ac{K19} is unfeasible. 
Another challenge is the relatively significant overlap between the complex stroma and the tumor class. The model tends to classify the tumor border area as complex stroma.

In Figure~\ref{fig:multisrc_cstr}, we display the visual results of the complex stroma detection on ROI 1 and 3, where desmoplastic reactions, and thus complex stroma, are present.
We show, from left to right, the reference images, the original ground truth labels, the extended ground truth labels with complex stroma, the DeepAll baseline, our previous \ac{SRA} work, and as well the results of the presented \ac{SRMA} model ($\mathcal{L}_{\textrm{IND}}^{1:1}$ and $\mathcal{L}_{\textrm{CRD}}^{K:1}$ setting). 
}

\ac{SRMA} outperforms the baselines in terms of pixel-wise accuracy, Jaccard index (\ac{IoU}), and Cohen's kappa score $\kappa$. 
Notably, the detection of the tumor is much more detailed compared to the single-source approach in both \ac{ROIs}.
Parts of the tissue previously considered as tumor can now be properly matched, thanks to the introduction of the complex stroma class.

Another interesting result in ROI 3 is that all the stromal areas are now considered as either complex stroma, tumor, or lymphocytes by all models.
This highlights how challenging the classification of complex stroma is without access to the higher-level context.
Pathologists also find this difficult, as they rely not only on the tissue morphology for this assessment but also on the spatial relations, i.e., the proximity to the tumor area.
Here, according to our extended ground truth, the complex stroma only surrounds the tumor region. 
However, the tissue tear disconnected some of the tumor surrounding regions, which suggests that the complex stroma area, in reality, spans even further.
This correlates with the prediction of both models, which identify the whole region as complex stroma.

\new{
Lastly, using the multi-source setting allows the introduction of a new class such as complex stroma to the detection task.
In the presented setting, the source domains do not need one-to-one class correspondences for the model to learn meaningful cross-domain features.
Here, \ac{CRCTP} does not include mucin, background, and adipose while \ac{K19} does not contain complex stroma. 
This is an interesting outcome, as it shows that new data that might even be acquired under different circumstances can be added with additional tissue classes without interfering with or altering the performance of the existing classes.
}

A visualization of the multi-source domain embedding space as well as the patch-based segmentation of a full \ac{WSI} image are available in \ref{app:tsne-multi}-\ref{app:wsi_classification}.

\section{Conclusion and Future Work}
\label{sec:conclusion}

In this work, we explore the usefulness of self-supervised learning and \ac{UDA} for the identification of histological tissue types. 
Motivated by the difficulty of obtaining expert annotations, we explore different \ac{UDA} models using a variety of label-scarce colorectal cancer histopathology datasets.

As our main contribution, we present a new label transferring approach from partially labeled, public datasets (source domain) to unlabeled target domains.
This is more practical than most previous \ac{UDA} approaches which are often tailored to fully annotated source domain data or tied to additional network branches dedicated to auxiliary tasks. 
Instead, we perform progressive cross-entropy minimization based on the similarity distribution among the unlabeled target and source domain samples, yielding discriminative and domain-agnostic features for domain adaptation.

In reality, not all tissue types are equally present in a \ac{WSI}, and some are quite rare.
Thus, the extracted patches are imbalanced in regards to class labels (categories), which imposes significant challenges for the trained models to generalize well.
For example, mucin is frequently present in mucinous carcinoma but is scarcely found in adenocarcinomas. 
Throughout various label transfer tasks, we show that our proposed \acf{SRMA} method can discover the relevant semantic information even in the presence of few labeled source samples, and yields a better generalization on different target domain datasets. 
Moreover, we show that our model definition can be generalized to a multi-source setting. As a result, the proposed model is able to learn rich data representation using multiple source domains.

Another example is the complex stroma class, which can be further divided into three subcategories (immature, intermediate, or mature), whose occurrences are highly variable and which are linked to patients' prognostic factor \citep{okuyama2020myxoid}.
Possible future work could take this class imbalance across \acp{WSI} into account and aim to improve the quality and variety of the provided positive and negative examples.

In addition, publicly available datasets are so far mostly composed of curated and thus homogeneous patches in terms of tissue types.
This data, however, do not capture the heterogeneity and complexity of patches extracted from images in the diagnostic routine. 
This can lead to erroneous detections, e.g., background and stroma interaction being interpreted as adipose tissue.
Thus, finding a self-supervised learning approach that can also properly embed mixed patches is a possible future extension of this work.

\new{Furthermore, the \ac{SRMA} framework is also highly modular and can thus be used for similar problems in other image analysis research fields. The selected backbone can be replaced, and the used data augmentations adapted to better fit with the task and data at hand.}

Lastly, the patch-based segmentation using our method can also be applied in a clinical context.
Many clinically relevant downstream tasks depend on accurate tissue segmentation, such as tumor-stroma ratio calculation, disease-free survival prediction, or adjuvant treatment decision-making.

\section*{Acknowledgments}
This work was supported by the Personalized Health and Related Technologies grant number 2018-327, and the Rising Tide foundation with the grant number CCR-18-130. The authors would like to thank Dr. Felix Müller for the annotation of the \ac{WSI} crops, M. Med. Philipp Zens for his feedback on the complex stroma detection, \new{Dr. Christophe Ecabert for providing GPU resources, and Guillaume Vray for the computation of the CycleGAN baseline,} which all greatly helped the evaluation of our method.

\bibliographystyle{unsrtnat}
\bibliography{main}  

\appendix

\section{Selection of Self-supervised Model}
\label{app:selfsupmodel}

To assess which self-supervised model we should use as the backbone for the \ac{UDA}, we compare the performances of several \ac{SOTA} self-supervised methods (SimCLR \citep{chen2020simple}, SupContrast \citep{khosla2020supervised}, and MoCoV2 \citep{chen2020improved}), as well as
the performance of the standard supervised learning approach when facing different levels of data availability. 
The results are presented in Table~\ref{tbl:clspercentage}. 
We report the performance of the single domain classification on \ac{K16} and \ac{K19}. The supervised approach uses ImageNet pre-trained weights. The self-supervised baselines are trained from scratch. 
After self-supervised training, we freeze the weights, add a linear classifier on top, and train it until convergence.  
For SupContrast \citep{khosla2020supervised} we jointly train the representation and the classification as described in the original paper.

We find that MoCoV2 \citep{chen2020improved} outperforms the two other \ac{SOTA} approaches. 
On \ac{K16}, the model gains up to $10\%$ in terms of the F1-score with respect to the other self-supervised baselines. 
In addition, MoCoV2 gives competitive results with the supervised baseline that is initialized with ImageNet weights. 
It shows that MoCoV2 is able to efficiently learn from unlabeled data and  create a generalized feature space.
This mainly comes from the combination of the momentum encoder and the access to a large number of negative samples.
Hence, we choose to adapt MoCoV2 for our proposed \ac{UDA} method.

\section{Patch Classification - t-SNE Projection}
\label{app:tsne}

\begin{figure*}[ht!]
\centering
  \includegraphics[width=\textwidth]{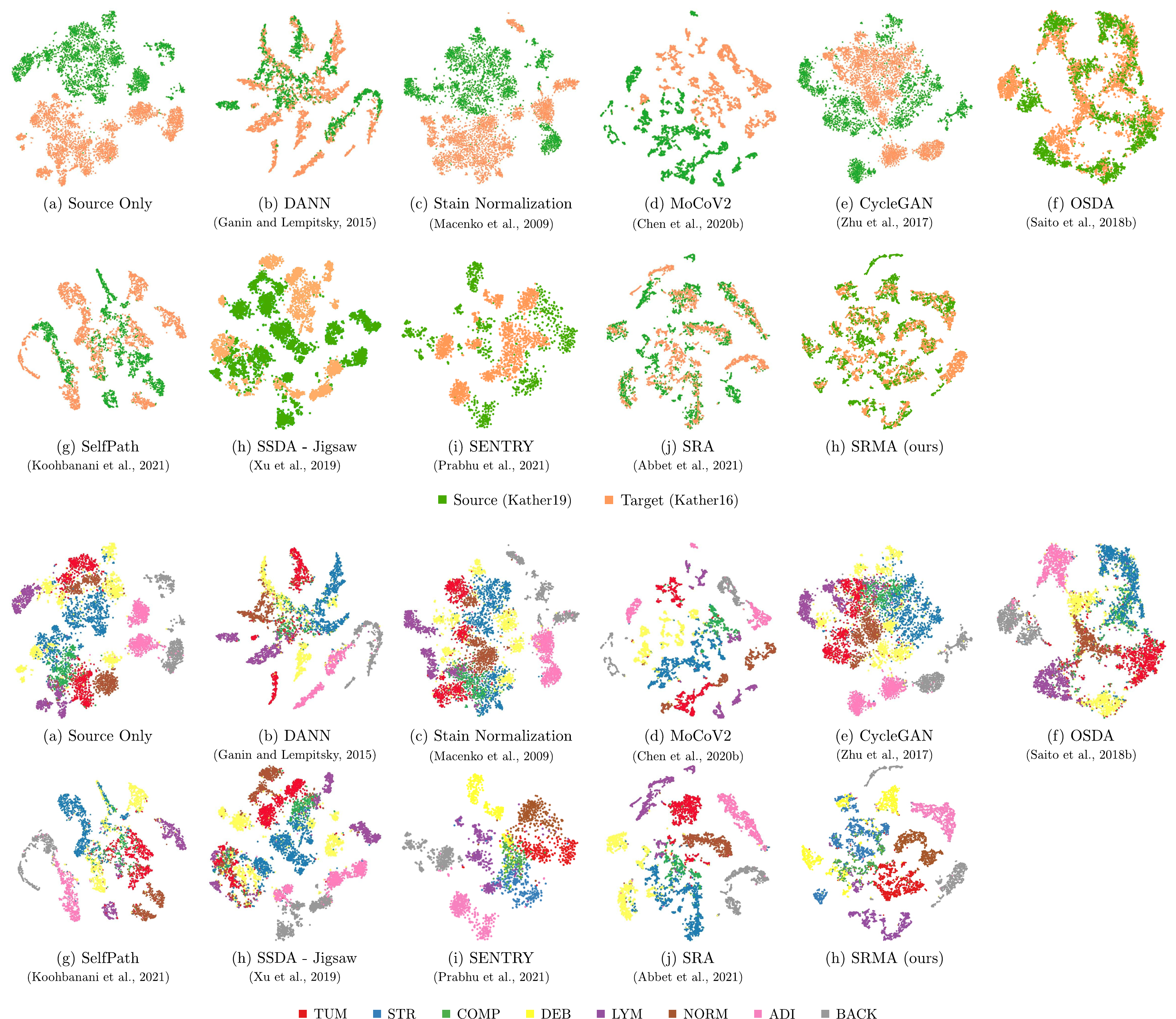}
  \caption{t-SNE projection of the source (\acl{K19}) and target (\acl{K16}) domain embeddings. We show the alignment of the embedding space as well as the individual classes for all presented models between the source and target domain. The classes of \acl{K19} are merged and relabeled according to the definitions in \acl{K16}. The standard supervised approach is depicted in (a). We compare our approach (i) to other domain adaptation methods (b-j). \new{Our approach (h) qualitatively show the best alignment between the source and target domain.}}
  \label{fig:sup_tsne_full}
\end{figure*}

In this section, we present the complementary results to the ones in Section~\ref{subsec:crossdomain_cls} for patch classification. The embeddings of all baselines and our proposed approach are displayed in Figure~\ref{fig:sup_tsne_full} using t-SNE visualization. We show the alignment between the source (\ac{K19}) and target (\ac{K16}) embedding domain, as well as classes-wise.

\new{With the source only approach, we can observe the lack of domain alignment between the feature spaces. Here, the model learns two distinct distributions for each set. On the other side, our approach shows a satisfactory alignment of domains compared to most baselines. The target complex stroma (K16) is linked to tumor, debris, lymphocytes, and stroma in the source domain (K19).}

\section{\new{Multi-Source Dataset Sampling Ratio}}
\label{app:bs_sampling}

\begin{table*}[ht]
    \caption{Classification results of the different \ac{SOTA} self-supervised approaches, as well as the supervised baseline on the \acf{K19} and \acf{K16} patch classification tasks. We present the results for different percentages of available training data. The top results are highlighted in bold. We report the weighted F1 score.}
    \label{tbl:clspercentage}
    \centering
    \begin{footnotesize}
    \begin{threeparttable}
    \begin{tabular}{l c c c l c c c}
    \toprule
    \toprule
        & \multicolumn{3}{c}{\acl{K16}} & & \multicolumn{3}{c}{\acl{K19}} \\
        & \multicolumn{3}{c}{Labels fraction} & & \multicolumn{3}{c}{Labels fraction} \\
    \cmidrule{2-4} 
    \cmidrule{6-8}
    Methods & $10\%$ & $20\%$ & $50\%$ & & $1\%$ & $2\%$ & $5\%$\\
    \midrule
    Supervised$^{\ddagger}$ & 85.8\tnote{**} & 86.5\tnote{**} & 87.9\tnote{**} &  & \textbf{89.2}\tnote{+} & \textbf{89.9}\tnote{+} & \textbf{90.5}\tnote{+}\\
    SimCLR \citep{chen2020simple} & 79.6\tnote{**} & 78.9\tnote{**} & 78.6\tnote{**} & & 76.9\tnote{**} & 79.4\tnote{**} & 80.7\tnote{**}  \\
    SupContrast \citep{khosla2020supervised} & 60.8\tnote{**} & 73.2\tnote{**} & 80.8\tnote{**} & & 78.7\tnote{**} & 81.6\tnote{**} & 85.0\tnote{**}\\
    MoCoV2 \citep{chen2020improved} & \textbf{88.5} & \textbf{90.2} & \textbf{91.1} & & \textbf{89.9} & \textbf{90.3} & \textbf{90.6} \\
    \bottomrule
    \bottomrule
    \end{tabular}
    \begin{tablenotes}
        \item[$\ddagger$] Model initialized with ImageNet pre-trained weights.
        \item[+] $\ p\geq0.05$; $^{*}\ p<0.05$; $^{**}\ p<0.001$; unpaired t-test with respect to the top result.
    \end{tablenotes}
    \end{threeparttable}
    \end{footnotesize}
\end{table*}

\begin{table*}[htb]
    \caption{\new{
    Study of the multi-source domain performance of the \acf{SRMA} approach with different sampling ratios. We use \ac{K19} and \ac{K16} as source datasets and \ac{CRCTP} as the target dataset. For \ac{K19} and \ac{K16}, only $1\%$ and $10\%$ of the source labels are used, respectively. For the proposed \ac{SRMA} model we compare the introduced multi-source approaches defined in Equations~\ref{eq:L_ind_11}-\ref{eq:L_crd_K1}, where $1:1$ and $K:1$ refers to the one-to-one and $K$-to-one setting, respectively. The probability of sampling an example from each set within a batch is indicated. We report the F1 score for the individual classes and weighted F1 score as the overall mean performance (all) averaged over 10 runs.
    }}
    \label{tbl:k16-k19-crctp-bs}
    \centering
    \scriptsize
    \begin{threeparttable}
    \begin{tabular}{lccccccccccrrrrrrrr}
    \toprule
    \toprule
    && \multicolumn{2}{c}{Multi-source} && \multicolumn{3}{c}{Sampling probability} \\
    \cmidrule{3-4}
    \cmidrule{6-8}
    Model && $\mathcal{L}_{\textrm{IND}}$ & $\mathcal{L}_{\textrm{CRD}}$ && K19 & K16 & CRCTP & Batch size && TUM & STR\tnote{$\ddagger$} & LYM & NORM & DEB & ALL\\
    \midrule
    \new{DeepAll \citep{dou2019domain}} && \new{-} & \new{-} && \new{-} & \new{-} & \new{-} & \new{128} && \new{72.4\tnote{**}} & \new{88.6\tnote{**}} & \new{43.6\tnote{**}} & \new{53.2\tnote{**}} & \new{71.8\tnote{**}} & \new{73.2\tnote{**}}
    \\
    \new{SRA\citep{abbet2021selfrule}} && \new{$1:1$} & \new{$1:1$} && \new{0.25} & \new{0.25} & \new{0.50} & \new{128} && \new{86.2\tnote{**}} & \new{87.6\tnote{**}} & \new{66.7\tnote{**}} & \new{71.0\tnote{**}} & \new{\textbf{80.5}} & \new{81.8\tnote{**}}
    \\
    \new{SRMA} && \new{$1:1$} & \new{$1:1$} && \new{0.25} & \new{0.25} & \new{0.50} & \new{128} && \new{\textbf{92.5}} & \new{88.4\tnote{**}} & \new{68.7\tnote{**}} & \new{68.3\tnote{**}} & \new{74.2\tnote{*}} & \new{82.9\tnote{*}}
    \\
    \new{SRMA} && \new{$K:1$} & \new{$1:1$} && \new{0.25} & \new{0.25} & \new{0.50} & \new{128} && \new{91.5\tnote{*}} & \new{87.6\tnote{**}} & \new{\textbf{70.7}} & \new{\textbf{75.0}} & \new{65.7\tnote{**}} & \new{82.7\tnote{*}}
    \\
    \new{SRMA} && \new{$1:1$} & \new{$K:1$} && \new{0.25} & \new{0.25} & \new{0.50} & \new{128} && \new{90.1\tnote{**}} & \new{\textbf{90.1}} & \new{\textbf{69.6}\tnote{+}} & \new{72.9\tnote{**}} & \new{71.6\tnote{**}} & \new{\textbf{83.6}} 
    \\
    \new{SRMA} && \new{$K:1$} & \new{$K:1$} && \new{0.25} & \new{0.25} & \new{0.50} & \new{128} && \new{\textbf{91.6}} & \new{87.4\tnote{**}} & \new{68.7\tnote{**}} & \new{73.9\tnote{**}} & \new{53.3\tnote{**}} & \new{81.2\tnote{**}}
    \\
    \midrule
    \new{SRMA} && \new{$1:1$} & \new{$1:1$} && \new{0.33} & \new{0.33} & \new{0.33} & \new{128} && \new{\textbf{92.9}\tnote{+}} & \new{87.8\tnote{**}} & \new{68.3\tnote{**}} & \new{65.3\tnote{**}} & \new{72.0\tnote{*}} & \new{82.0\tnote{**}}   
    \\
    \new{SRMA} && \new{$K:1$} & \new{$1:1$} && \new{0.33} & \new{0.33} & \new{0.33} & \new{128}  && \new{\textbf{93.1}} & \new{87.3\tnote{**}} & \new{70.5\tnote{**}} & \new{\textbf{78.3}} & \new{66.9\tnote{**}} & \new{83.4\tnote{*}}    
    \\
    \new{SRMA} && \new{$1:1$} & \new{$K:1$} && \new{0.33} & \new{0.33} & \new{0.33} & \new{128} && \new{92.5\tnote{*}} & \new{\textbf{89.7}} & \new{\textbf{71.6}} & \new{73.0\tnote{**}} & \new{66.9\tnote{**}} & \new{\textbf{83.8}}      
    \\
    \new{SRMA} && \new{$K:1$} & \new{$K:1$} && \new{0.33} & \new{0.33} & \new{0.33} & \new{128} && \new{92.2\tnote{*}} & \new{88.6\tnote{**}} & \new{66.1\tnote{**}} & \new{74.3\tnote{**}} & \new{\textbf{74.5}} & \new{83.4\tnote{*}}    
    \\
    \midrule
    \new{SRMA} && \new{$1:1$} & \new{$1:1$} && \new{0.40} & \new{0.20} & \new{0.40} & \new{128} && \new{90.5\tnote{**}} & \new{88.3\tnote{**}} & \new{63.8\tnote{**}} & \new{71.8\tnote{**}} & \new{66.1\tnote{**}} & \new{81.5\tnote{**}}       
    \\
    \new{SRMA} && \new{$K:1$} & \new{$1:1$} && \new{0.40} & \new{0.20} & \new{0.40} & \new{128} && \new{90.8\tnote{**}} & \new{\textbf{89.8}} & \new{62.0\tnote{**}} & \new{\textbf{74.7}} & \new{64.1\tnote{**}} & \new{82.2\tnote{**}}       
    \\
    \new{SRMA} && \new{$1:1$} & \new{$K:1$} && \new{0.40} & \new{0.20} & \new{0.40} & \new{128} && \new{92.0\tnote{*}} & \new{88.6\tnote{**}} & \new{\textbf{69.5}} & \new{73.7\tnote{**}} & \new{64.8\tnote{**}} & \new{82.8\tnote{**}}       
    \\
    \new{SRMA} && \new{$K:1$} & \new{$K:1$} && \new{0.40} & \new{0.20} & \new{0.40} & \new{128} && \new{\textbf{92.7}} & \new{89.3\tnote{**}} & \new{65.8\tnote{**}} & \new{\textbf{74.7}\tnote{+}} & \new{\textbf{75.2}} & \new{\textbf{83.8}}       
    \\
    \bottomrule
    \bottomrule
    \end{tabular}
    \begin{tablenotes}
        \item[$\ddagger$] \new{The STR and MUS classes are merged as STR class; DEB and MUC classes as DEB.}
        \item[+] $\ p\geq0.05$; $^{*}\ p<0.05$; $^{**}\ p<0.001$; unpaired t-test with respect to top
    \end{tablenotes}
    \end{threeparttable}
\end{table*}

\new{
When performing multi-source domain adaptation, we assume that the distribution of all the source and target samples are the same. More formally, we have $p(\textbf{x} \in \mathcal{D}_s) = K p(\textbf{x} \in \mathcal{D}_s^k) = p(\textbf{x} \in \mathcal{D}_t)$. This section, we analyze the importance of balancing the source and target domains during the pre-training stage. We use \ac{K19} and \ac{K16} as source datasets and \ac{CRCTP} the target dataset. For \ac{K19} and \ac{K16}, only $1\%$ and $10\%$ of the source labels are used, respectively.

The results of the classification performance on the \ac{CRCTP} dataset are presented in Table~\ref{tbl:k16-k19-crctp-bs}. We indicate the multi-source scenario ($1:1$ or $K:1$), the sampling probability for each of the dataset, and the batch size.

The cross-domain matching using the $K:1$ scenario shows the highest variance and its performances can change up to $2.6\%$. Overall, we can observe that balanced probability between all sets, namely $\frac{1}{3}$ each, gives similar results across all multi-source scenarios. In addition, when lowering the sampling probability of \ac{K16} we can see a drop in performances. This suggests that it is important to have a balanced sampling strategy even if one of the source sets (e.i., \ac{K16} with 5,000 examples) is much smaller.
}

\section{Multi-source - t-SNE Projection}
\label{app:tsne-multi}

\begin{figure*}[htb]
\centering
  \includegraphics[width=0.93\textwidth]{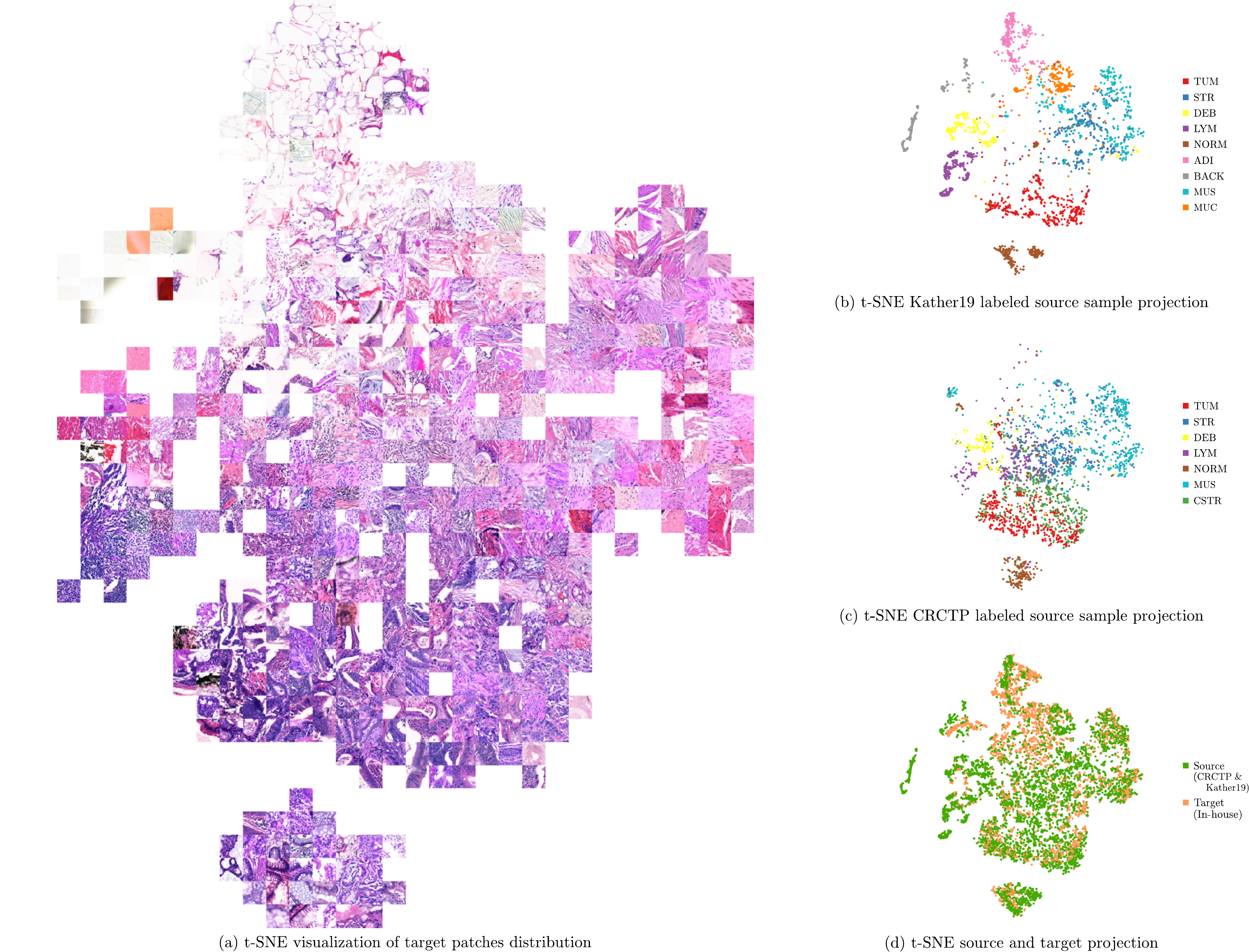}
  \caption{
  t-SNE visualization of the \ac{SRMA} model trained on \ac{CRCTP}, \ac{K19} and the in-house dataset. All sub-figures depict the same embedding. 
(a) Patch-based visualization of the embedding.
  (b-c) Distribution of the labeled source samples. 
  (d) Relative alignment of the source and target domain samples.
  }
  \label{fig:crctp_k19_bern_tsne_img}
\end{figure*}

Figure~\ref{fig:crctp_k19_bern_tsne_img} shows the visualization of the embedding for the proposed multi-source domain adaptation in Sections~\ref{subsec:multi_source_wsi}. It highlights the alignment of the feature space between the two source sets (\ac{K19}, \ac{CRCTP}) and our in-house dataset.

\new{
We observe that for each source domain, the categories are well clustered. Moreover, we notice that the classes shared by both domains (e.i., namely tumor, stroma, debris, lymphocytes, normal mucosa, and muscle) fully overlap. In addition, the tissues that are domain-specific (e.i., adipose, background, mucin, and complex stroma) form individual groups. Subsequently, it indicates that our approach was able to properly correlate similar tissue definitions across the source domains while maintaining domain-specific tissue representation.
}

\new{
Looking at the source and target projection, we discern a batch of tissue (center-top) that does not align with the source domain. When associated with the patches visualization, we can recognize tiles that include loose stroma, collagen, or blood vessels representation. Rightfully, none of the mentioned classes were present in the source domain, thus proving the usefulness of the easy-to-hard approach.
}




\section{\new{Patch-based Segmentation of \ac{WSI} from the TCGA cohort}}
\label{app:wsi_classification}

\new{
In this section we highlight the performance of our framework on a publicly available \ac{WSI} (UUDI: 2d961af6-9f08-4db7-92b2-52b2380cd022) from the TCGA colon cohort \citep{coad2016, read2016}. 
We apply our trained \ac{SRMA} framework, as described in section \ref{subsec:crossdomain_seg}, where \ac{K19} is used as the source domain and our in-house domain as the target one. 
We show the original image, as well as the classification output and the tumor class probability map of our proposed \ac{SRMA} method. 

The model is able to accurately classify tissue across the whole slide. Moreover, the pipeline gives a rather detailed output which is a remarkable performance for a patch-based approach that was not specifically designed for segmentation purposes. Moreover, the model is agnostic to artifacts such as the permanent marker spots (green marks on the bottom left). The tumor prediction map gives an overview of the tumor class probability across the \ac{WSI}. This class is of particular interest, as tumor detection is an important step for many downstream tasks, e.g., detection of the invasive front or the tumor stroma ratio.}

\begin{figure*}[!ht]
    \centering
    \includegraphics[width=0.9\textwidth]{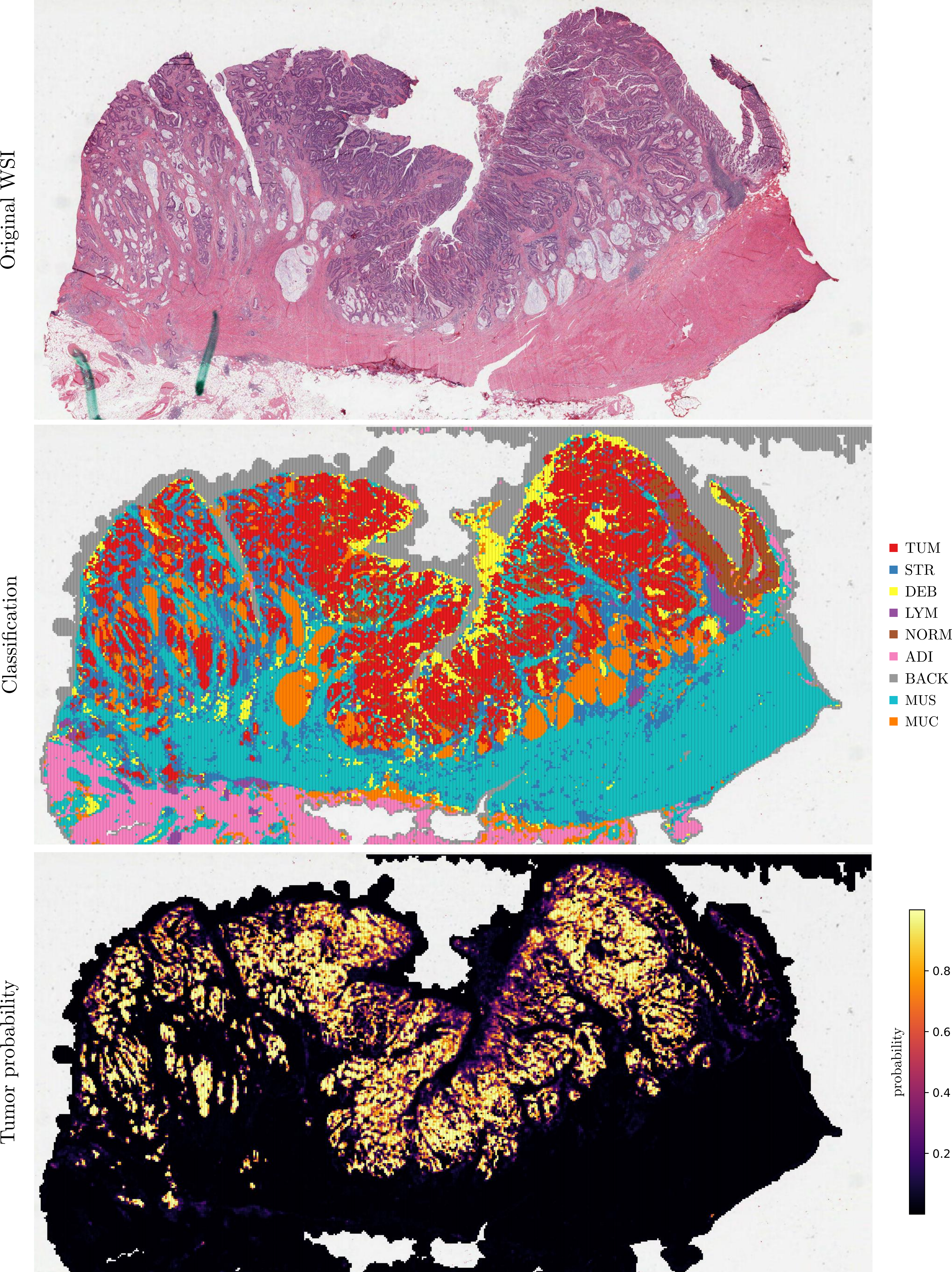}
    \caption{\new{Segmentation results on a sample \ac{WSI} from the TCGA cohort achieved by our \ac{SRMA} model trained using \ac{K19} as the source dataset and our in-house set as the target dataset. From top to bottom, we show the original image, the classification output, and tumor class probability map.}}
    \label{fig:wsi-overlay}
\end{figure*}

\end{document}